%% file: acl_latex.tex
\pdfoutput=1

\documentclass[11pt]{article}

\usepackage[preprint]{acl}

\usepackage{times}
\usepackage{latexsym}

\usepackage[T1]{fontenc}

\usepackage[utf8]{inputenc}

\usepackage{microtype}

\usepackage{inconsolata}
\usepackage{graphicx}
\title{Efficiently Quantifying and Mitigating Ripple Effects in Model Editing}

\author{Jianchen Wang, Zhouhong Gu, Xiaoxuan Zhu, Lin Zhang, Haoning Ye, Zhuozhi Xiong,\\ \textbf{Hongwei Feng, Yanghua Xiao} \\
Fudan University \\
}

\usepackage[utf8]{inputenc} 
\usepackage[T1]{fontenc}    
\usepackage{hyperref}       
\usepackage{url}            
\usepackage{booktabs}       
\usepackage{amsfonts}       
\usepackage{nicefrac}       
\usepackage{microtype}      
\usepackage{amsmath}
\usepackage{wrapfig}
\usepackage{color,xcolor}

\usepackage{tikz}
\input{latexs/include}

\begin{document}

\maketitle
\begin{abstract}
\input{latexs/00Abstract}
\end{abstract}

\section{Introduction}
\input{latexs/01Introduction.14}

\section{Related Work}
\input{latexs/02RelatedWork.6}

\section{Preliminary}
\input{latexs/03Preliminary.7}

\section{Our Method}
\input{latexs/04OurMethod.8}

\section{Experiment Setup}
\input{latexs/05ExperimentSetup.3}

\section{Experiment}

\input{latexs/06Experiment.4}
\section{Conclusion}
\input{latexs/07Conclusion.2}

\section*{Limitation}
\input{latexs/Limitation}

\section*{Ethics Statement}

Model editing involves changing how language models output. Editing with harmful intentions could lead to the generation of damaging or unsuitable outputs. Therefore, it's essential to ensure safe and harmless model editing. Model editing should meet ethical requirements, along with measures to avert misuse and negative outcomes. Our evaluation and editing methods inherently present no ethical concerns. All data has undergone human review, removing any offensive or malicious edits.

\bibliography{custom}

\appendix

\section{Appendix}
\label{sec:appendix}
\input{latexs/09Appendix.1}

\end{document}

%% file: latexs/include.tex
\usepackage{times}
\usepackage{textcomp}
\usepackage{latexsym}
\usepackage{colortbl}
\usepackage{CJK}
\definecolor{mygray}{gray}{0.8}

\usepackage{amsfonts,amssymb,bbm}
\usepackage{arydshln}
\usepackage{hhline} 
\usepackage{booktabs} 
\usepackage{makecell} 
\usepackage{multirow}
\usepackage[normalem]{ulem}
\useunder{\uline}{\ul}{}
\usepackage{tabularx}

\usepackage{pifont}
\newcommand{\cmark}{\ding{51}}%
\newcommand{\xmark}{\ding{55}}%

\usepackage{amsmath}
\usepackage[T1]{fontenc}

\usepackage[utf8]{inputenc}

\usepackage{microtype}

\usepackage{inconsolata}

\usepackage{graphicx} 
\usepackage{mathrsfs}
\usepackage{cancel}

%% file: latexs/00Abstract.tex
Large Language Models have revolutionized numerous tasks with their remarkable efficacy. However, editing these models, crucial for rectifying outdated or erroneous information, often leads to a complex issue known as the ripple effect in the hidden space. While difficult to detect, this effect can significantly impede the efficacy of model editing tasks and deteriorate model performance. This paper addresses this scientific challenge by proposing a novel evaluation methodology, Graphical Impact Evaluation(GIE), which quantitatively evaluates the adaptations of the model and the subsequent impact of editing. Furthermore, we introduce the Selective Impact Revision(SIR), a model editing method designed to mitigate this ripple effect. Our comprehensive evaluations reveal that the ripple effect in the hidden space is a significant issue in all current model editing methods. However, our proposed methods, GIE and SIR, effectively identify and alleviate this issue, contributing to the advancement of LLM editing techniques.

%% file: latexs/01Introduction.14.tex
The rapid progress of Large Language Models (LLMs) has demonstrated remarkable effectiveness across a wide range of tasks~\cite{brown2020language,zhao2023survey,openai2023gpt4,touvron2023llama,gu2023xiezhi}. 
However, many facts embedded within these models may need to be updated or contain errors~\cite{lazaridou2021mind,dhingra2022time, jang2022towards}.
As a result, methods for editing these facts within LLMs have gained increasing attention~\cite{zhu2020modifying,de-cao-etal-2021-editing,meng2022locating,meng2023massediting,si2023prompting}.
The primary goal of model editing is to refine the factual memory of LLMs in specific domains, ensuring targeted improvements without compromising overall factual memorization accuracy.
This process requires a delicate balance between successfully implementing factual edits and preventing unintended damage to the model's memorization of other facts.

Despite the effectiveness of many model editing techniques in various situations, studies have revealed that model editing harms the LLMs' memory of other facts, a phenomenon known as the ``ripple effect''~\cite{gu2024model}.
The ripple effect is categorized into two primary types by previous research:
``Ripple Effect in the Same Entity'' and ``Ripple Effect in Hidden Space''.
The former occurs when editing knowledge about an entity potentially damages the model's memory of other facts related to that entity~\cite{li2023unveiling,yao2023editing}.
The latter arises when changing the model's memory of an entity in a hidden space affects other entities close to it in that space~\cite{hoelscherobermaier2023detecting,sakarvadia2023memory}.

However, we argue that the ``Ripple Effect in the Same Entity'' is inherently encompassed by the ``Ripple Effect in Hidden Space''.
When model editing induces changes in the model's parameters, it inevitably alters the knowledge influenced by those parameters, leading to broader, unintended modifications.
A key challenge posed by the ripple effect in hidden space is its elusive nature. It lacks direct factual links to the edited object, making it difficult to detect and address the implicit impact on seemingly unrelated entities.
As the number of edits increases, the failure to mitigate this hidden ripple effect leads to a steep deterioration in model performance, ultimately rendering the edited models unreliable and potentially harmful when applied in real-world scenarios~\cite{li2023unveiling,wang2023knowledge}.
Therefore, detecting and controlling the ripple effect in hidden space is essential to ensure the reliability, stability, and practicality of model editing techniques.

We first introduce a novel quantitative evaluation method called Graphical Impact Evaluation (GIE) to address this challenge. 
Specifically, GIE selects edit targets from Knowledge Graphs (KGs), which typically contain many facts, and evaluates the most significantly affected factual knowledge based on the differences in edit targets. This design stems from one of our findings, which indicates that model editing preferentially impacts other facts with embeddings similar to the edited facts.
By evaluating the model's changes in response to these most easily influenced facts, GIE effectively and efficiently assesses the Ripple Effect in hidden space.

Building upon the concept of GIE, we further propose an efficient and effective method to mitigate the ripple effects, named Selective Impact Revision (SIR). 
SIR suppresses the ripple effects of model editing by selecting and retraining facts in the KG that are closely related to the edited facts during the model editing process.
By focusing on the most relevant facts identified through GIE, SIR efficiently targets the root cause of the ripple effect and minimizes its impact on the model's performance.

The GIE method revealed that even the state-of-the-art (SOTA) model editing approach is significantly impacted by ripple effects in the latent space, with 16.51\% of unrelated facts experiencing severe consequences.
The SIR method demonstrated a 54.75\% reduction in the the ripple effect intensity within the hidden space compared to the SOTA model editing technique.



%% file: latexs/02RelatedWork.6.tex
\subsection{Knowledge Editing}





The knowledge Model Editing method is essential for incorporating new knowledge into existing LLM while maintaining the integrity of pre-existing information.
These techniques are generally grouped into three primary categories.
The first is external memorization-based methods, which involve the use of separate memory modules to store new knowledge, thus leaving the original model's weights unchanged, offering scalability and the possibility to expand knowledge without altering the structure of the pre-trained model\cite{li2022large,madaan2022memory,mitchell2022memory,murty2022fixing}.
The second category is global optimization-based methods, which consist of extensive updates across the model, influenced by newly acquired knowledge, which, although ensuring comprehensive modification, can be resource-demanding due to the extensive parameter space\cite{sinitsin2019editable,de-cao-etal-2021-editing,hase2021language,mitchell2022fast,gangadhar2024model}.
Last is local modification-based methods focus on adjusting specific parameters, providing a targeted and more resource-efficient means of integrating new knowledge into LLMs\cite{dai2022knowledge,li2023pmet,meng2022locating,meng2023massediting}\cite{wang2023knowledge}.
This paper primarily focuses on Global Optimization-based Methods and Local Modification-based Methods, both of which involve updating the model. We also experiment with the latest method ICE~\cite{cohen2023evaluating, chen2024robust, mitchell2022memory}.
We aim to address the challenges associated with these methods, particularly the ripple effect in the hidden space, which has yet to be largely overlooked in previous research.


\subsection{Knowledge Editing Evaluation}
There has been an increasing focus on the evaluation of model editing. 
The primary benchmarks currently employed to assess editing methods are Zero-Shot Relation Extraction(zsRE) \citep{levy-etal-2017-zero} and CounterFact \citep{meng2022locating}.
zsRE is a question-answering dataset designed for relation-specific queries. It is annotated with human-generated question paraphrases that can measure the model's robustness to semantically equivalent inputs. 
CounterFact is a more challenging evaluation dataset that introduces counterfactual edits. 
RippleEdits~\cite{cohen2023evaluating} is a benchmark evaluating the ``ripple effects'' in knowledge editing. 
Specifically, one should go beyond the single edited fact and check that other facts logically derived from the edit were also changed accordingly. 
In addition, research \cite{hoelscher-obermaier-etal-2023-detecting,li2023unveiling} shows that existing editing methods can have unwanted side effects on LLMs. 
This paper primarily focuses on these unwanted side effects, a topic not thoroughly explored in previous studies.
Unlike previous work, we posit that the primary source of the Ripple Effect stems from latent space correlations, with other types of Ripple Effects being derivatives of this underlying relationship. Our research investigates how knowledge graphs can help uncover the extent of these side effects and highlight the discrepancies in knowledge distribution between models and human understanding. This study significantly advances how model editing can implicitly impact other knowledge within the model.

%% file: latexs/03Preliminary.7.tex
\textbf{Knowledge Graph (KG}), represents as $S$ in this paper, is a large-scale semantic network that uses collections of triplets to include all kinds of factual knowledge.
Triplets $S=\{\langle s, r, o\rangle\}$, the fundamental unit of knowledge representation in a KG, typically consists of a subject, relation, and object, representing either the relationship between entities or an attribute value of an entity.

\textbf{Model Editing} is a method that focuses on applying factual updates to LMs.
This approach involves converting an edit target, represented as a triple $\langle s_e, r_e, o_e\rangle$, into a free-text prompt.
The existing LM, denoted as $f_\theta$, is then fine-tuned using this prompt to incorporate the new factual information while maintaining its pre-existing knowledge and capabilities~\cite{cohen2023evaluating}.

\textbf{Factual Change} refers to the overall changing of the LM's memorization of factual knowledge.
Given a fact set $\mathcal{F}$ and a corresponding set of changes $\Delta \mathcal{F} (|\Delta \mathcal{F}| \leq |\mathcal{F}|)$, the post-change fact set in LM is expressed as
\vspace{-3mm}
\begin{equation}
\mathcal{F'} = \mathcal{F} + \Delta \mathcal{F} + R(\Delta \mathcal{F}),
\label{eq:factual_change}
\vspace{-3mm}
\end{equation}
where $R(\Delta \mathcal{F})$ signifies the ripple effect induced by $\Delta \mathcal{F}\vspace{2mm}$.

\textbf{Ripple Effect}, a side effect of model editing, arises from modifications to a language model's memory of specific factual knowledge, causing changes in the model's internal parameters and consequently impacting its memory of other factual knowledge.
Model editing methods aim to eliminate this ripple effect completely.





%% file: latexs/04OurMethod.8.tex
\subsection{Graphical Impact Evaluation~(GIE)}

\textbf{Naive Ripple Effect Evaluation Method:}
The ripple effect $R(\Delta \mathcal{F})$ introduced by model editing can be quantified by measuring the change in the evaluation metric on all fact memory that is not the edited target within the edited LLM, computed by the post-edit model \(f_{\theta_e}\) and the pre-edit model \(f_\theta\):
\begin{equation}
R(\Delta \mathcal{F}) = \text{Metric}[f_{\theta_e}(\mathcal{F'}\setminus\Delta \mathcal{F})] - \text{Metric}[f_\theta(\mathcal{F})].
\label{eq:ripple_effect_on_facts}
\end{equation}
However, obtaining all factual knowledge is extremely challenging. 
Directly using existing KGs to evaluate the Ripple Effect quantitatively is an alternative~\cite{cohen2023evaluating}.
Existing KGs $S=\{\langle s, r, o\rangle\}$ contain vast factual knowledge and have undergone manual or automated validation to ensure quality. 
It also provides a standardized testing benchmark for evaluating various model editing methods.
In Eq.~\ref{eq:factual_change} and Eq.~\ref{eq:ripple_effect_on_facts}, $\mathcal{F'}$ is an ideal value, which is difficult to obtain. 
Therefore, the following approximation is often applied:
\begin{align}
&f_{\theta_{e}}(\mathcal{F'}\setminus\Delta\mathcal{F})\approx f_{\theta_{e}}(\mathcal{F});\notag\\ &S\approx \mathcal{F}; \Delta S\approx \Delta\mathcal{F}
\end{align}
After applying the above approximation, the evaluation of the Ripple Effect is reformulated from Eq.~\ref{eq:ripple_effect_on_facts} into the following format:
\begin{equation}
\begin{aligned}
\small
R(\Delta \mathcal{F}) = &\frac{1}{|S|}
  \sum_{\langle s, r, o \rangle \in S} (\text{Metric}[f_{\theta_e}(\langle s, r, o\rangle )] - \\
  &\text{Metric}[f_\theta(\langle s, r, o\rangle )]),
\end{aligned}
\end{equation}

\textbf{GIE Ripple Effect Evaluation Method:}
However, directly using the entire KG is precise yet highly inefficient.
GIE proposes to assess the metric changing in the triplets most similar to the edit targets to evaluate the ripple effect efficiently.
By quantifying the degree of change in these closely related triplets, the impact of the model editing can be effectively yet efficiently assessed, thereby measuring the Ripple Effect:
\begin{align}
\small 
S_{\text{selected}} =& \{ \langle s, r, o \rangle \mid \text{sim}(f_{\theta}\big(\langle s_e, r_e, o_e \rangle,\big.\notag\\ 
& f_{\theta}\big(\langle s, r, o \rangle\big)\big) > \tau)\}\big.\notag\\
\text{s.t.} \big(\langle &s_e, r_e, o_e \rangle\big)
\in \Delta \mathcal{S}, \big(\langle s, r, o \rangle \in \mathcal{S} \big),
\end{align}
\begin{align}
\small 
R(\Delta \mathcal{F}) & = \frac{1}{|S_{\text{selected}}|}
  \sum_{\langle s, r, o \rangle \in S_{\text{selected}}} \notag\\
  \big( \text{Metric}&[f_{\theta_e}(\langle s, r, o \rangle)] - \text{Metric}[f_\theta(\langle s, r, o \rangle)], \big)
\end{align}
$\text{sim}(\cdot,\cdot)$ is an embedding-based similarity function, and $\tau$ is a threshold defining the minimal similarity for inclusion in the evaluation set $S_{\text{selected}}$.

\subsection{Selective Impact Revision~(SIR)}

\textbf{Naive Ripple Effect Alleviation Method:}
The simplest method to mitigate the Ripple Effect is to train LMs with all facts memorized by the models alongside the specified edit target.
\begin{equation}
\min_{\theta} \sum_{f \in (\mathcal{F} \cup \Delta\mathcal{F})} \mathcal{L}(f; \theta),
\end{equation}
where \( \mathcal{L} \) is the loss function and \( \theta \) represents the model parameters. This approach aims to preserve the memory of all other facts while accommodating the edited facts.

However, as discussed in the previous subsection, obtaining all of an LM's factual memories is extremely challenging.
Therefore, directly using existing comprehensive KGs $S$ as a surrogate of all facts is a practical compromise:
\begin{equation}
\min_{\theta} \sum_{ \langle s,r,o\rangle \in (\mathcal{S} \cup \Delta\mathcal{S})} \mathcal{L}(\langle s,r,o \rangle; \theta).
\end{equation}
$\Delta\mathcal{S}$ here represents the edit targets.

\textbf{SIR Ripple Effect Alleviation Method:}
SIR proposes a more efficient approach by selectively retraining based on the metric changing between the edited and pre-edited LMs regarding the facts in the KGs.
For an edited model \( f_{\theta_e} \), the fact that triplets that suffer the most from the ripple effect are detected by the GIE method. 
SIR samples the top-K facts with the largest metric changing and retrains these facts.
\begin{equation}
\begin{aligned}
\small
S_{\text{selected}} =&\Big\{ \langle s, r, o \rangle \mid \text{Top}_{\text{K}}\Big( \text{Metric}[f_{\theta_e}(\langle s, r,\\
&o \rangle)] - \text{Metric}[f_\theta(\langle s, r, o \rangle)] \Big) \Big\},\\
&\text{s.t.} \big(\langle s, r, o\rangle \in \mathcal{F}\big),
\end{aligned}
\end{equation}
$\text{Top}_{\text{K}}(X)$ here represent the largest K number in $X$.
Moreover, the SIR training objective can be formulated as follows:
\begin{equation}
\min_{\theta} \sum_{\langle s, r, o\rangle \in \text{S}_{\text{selected}}} \mathcal{L}(\langle s, r, o\rangle; \theta),
\end{equation}







%% file: latexs/05ExperimentSetup.3.tex
\input{tables/simple_table}

The experiments are designed to address two questions: 
\textit{1)} Is there a method to efficiently and effectively quantify the ``ripple effect''? 
\textit{2)} Can ``ripple effect'' be effectively yet efficiently mitigated?

\subsection{Evaluation Dataset Construction}

\begin{table}
    \small
    \centering
    \resizebox{\columnwidth}{!}{
    \begin{tabular}{ccccc}
    \toprule
        \textbf{Name} & \textbf{\#Triplets} & \textbf{\#Entities} & \textbf{\#Relation} & \textbf{\#Prompt}\\
    \hline
        wiki5m  & 21,354,359 & 4,813,490 & 824 & - \\
        wiki30t & 30,319 & 10,571 & 269 & 14,148\\

    \bottomrule
    \end{tabular}}
    \caption{The statistic information to the KGs used in the experiments.}
    \label{tab:statistic}
    \vspace{-7mm}
\end{table}

GIE employs comprehensive KGs to assess the ripple effect, rather than conventional benchmarks such as COUNTERFACT~\cite{meng2022locating}, zsRE~\cite{levy-etal-2017-zero}, and RIPPLEEDITS~\cite{cohen2023evaluating}, which consists of a limited number of fixed prompts.
This limited scope resulted in the omission of the assessment of the ripple effect on broader facts.

However, given the vast scale of most KGs, which often contain billions of triplets, utilizing entire KGs for evaluation incurs prohibitive computational costs.
Therefore, the experimental analysis in this paper focuses on a subset of Wiki5m~\cite{wang2021kepler}.
The detailed statistic of the data we used in the experiment is listed in Tab.~\ref{tab:statistic}, and the specific experimental steps are as follows:

\textbf{Step 1: Subgraph Collection}
A Breadth-First Search (BFS) sampling method is employed to derive a representative subgraph from Wiki5m~\cite{wang2021kepler}.
This technique sequentially visits all entities that have relations with each other, resulting in a subnetwork called wiki30t that is closely connected.
The statistic information of Wiki5m and Wiki30t is listed in Tab.~\ref{tab:statistic}.

\textbf{Step 2: Prompt Generation}
Natural language prompts for each triplet are generated automatically using GPT-4o, ensuring consistency and fluency across the dataset.

\textbf{Step 3: Edit Target Selection}
The choice of edit targets varies, with different selection methods resulting in distinct distributions.
Using \textbf{BFS Sampling} results in highly concentrated edit targets, while \textbf{Random Sampling} produces more dispersed targets.
Each target must maintain a plausible degree of factual integrity (``The Eiffel Tower is located in Donald Trump'' is not a good edit fact, for example).
For each triplet $\langle s, r, o\rangle$, the edit target is modified to $\langle s, r, o'\rangle$, where $o'$ is chosen to maintain the same relation $r$ as the original object $o$, ensuring the edit remains plausible.

\subsection{Baseline}

\subsubsection{Ripple Effect Evaluation Method}

\textbf{Vanilla}: This evaluation focuses on the \textbf{neighbors of edited nodes} within knowledge graphs (KGs) to assess the ripple effects resulting from model edits. By quantifying changes in metrics among triplets that share factual relationships with the edited target, this approach effectively captures the extent of ripple effects caused by the modifications.
\textbf{GIE}: This method constructs a similarity graph based on the \textbf{semantic similarity between individual triplets} to evaluate the ripple effects induced by model edits. The GIE method is incredibly proficient at revealing how edits may impact nodes and connections that appear unrelated at first glance.

\subsubsection{Model Editing Method}
Here, we put the result of training-based baselines in Tab.~\ref{tab:simple_experiment}.
As for more baselines, which include outside storer-based methods, please refer to Tab.~\ref{tab:experiment} in the Appendix.
\textbf{Fine-tuning~(FT)} the model's parameters in a specific layer are updated using gradient descent with Adam optimizer and early stop strategy.
\textbf{Constrained Fine-Tuning(FT+L)}~\cite{zhu2020modifying} fine-tuning with an $L_{\infty}$ norm constraint on weight changes. 
\textbf{MEND}~\cite{mitchell2022fast} The model's parameters are updated through a hypernetwork using a low-rank gradient decomposition from standard fine-tuning.
\textbf{ROME}~\cite{meng2022locating} uses causal intervention for identifying neuron activations that are decisive in a model's factual predictions, then computes and inserts key-value pairs into specific MLP layers.
\textbf{MEMIT}~\cite{meng2023massediting} improves ROME for mass editing of diverse knowledge. For multiple edits, updates are distributed across various MLP layers in a top-down approach to avoid unintended impacts of inadvertently influencing edited layers when editing layers.
\textbf{SIR} represents our proposed methodology. 
SIR incorporates identifying and selective re-editing triplets for more effective and efficient model editing.
Additional implementation details are offered in Appendix~\ref{sec:implementation}.

\subsection{Metric}
We employ perplexity as the primary metric to measure the model's confidence in generating, for it is sensitive to shifts in the probability distribution.
If we have a tokenized sequence $X = \left(x_0, x_1, \dots, x_t \right)$, then the perplexity of $X$ is
\begin{equation}
    \text{\textit{PPL}}\left(X\right) = \exp \left\{- \frac{1}{t} \sum_{i}^{t} \log p_\theta\left(x_i \mid x_{<i} \right)  \right\},
\end{equation}
where $\log p_\theta\left(x_i \mid x_{<i} \right)$ is the log-likelihood of the $i$th token conditioned on the preceding $x_{<i}$ according to the model. 
Additional experiments utilizing alternative metrics~(BLEU, ROUGE) are documented in Appendix~\ref{sec:appendix2}.

%% file: tables/simple_table.tex
\begin{table*}[t]
\centering

\resizebox{\textwidth}{!}{
\begin{tabular}{cc|ccccccccl|ccccccccl}
\hline
&&\multicolumn{9}{c|}{\textbf{BFS Sampling}}&\multicolumn{9}{c}{\textbf{Random Sampling}}\\\cline{3-20} 
&&\multicolumn{3}{c}{1}&\multicolumn{3}{c}{2}&\multicolumn{3}{c|}{inf}&\multicolumn{3}{c}{1}&\multicolumn{3}{c}{2}&\multicolumn{3}{c}{inf}\\\cline{3-20} 
Methods&\#Edition&Vanilla&GIE&Diff&Vanilla&GIE&Diff&Vanilla&GIE&\multicolumn{1}{c|}{Diff}&Vanilla&GIE&Diff&Vanilla&GIE&Diff&Vanilla&GIE&\multicolumn{1}{c}{Diff}\\
\hline
\multicolumn{1}{c|}{\multirow{3}{*}{FT}}&\multicolumn{1}{c|}{1}&5.77&\textbf{10.99}&\multicolumn{1}{c|}{{\ul 5.22}}&9.35&\textbf{8.97}&\multicolumn{1}{c|}{-0.38}&\multicolumn{1}{l}{10.54}&\multicolumn{1}{l}{8.94}&{\ul -1.60}&-7.09&&\multicolumn{1}{c|}{}&0.92&&\multicolumn{1}{c|}{}&\multicolumn{1}{l}{5.07}&\multicolumn{1}{l}{0.44}&{\ul -4.63}\\
\multicolumn{1}{c|}{}&\multicolumn{1}{c|}{50}&7.17&\textbf{4.65}&\multicolumn{1}{c|}{-2.52}&4.78&\textbf{3.89}&\multicolumn{1}{c|}{-0.89}&3.73&5.23&1.50&3.21&\textbf{1.48}&\multicolumn{1}{c|}{-1.73}&1.92&\textbf{4.35}&\multicolumn{1}{c|}{2.43}&22.64&2.82&{\ul -19.82}\\
\multicolumn{1}{c|}{}&\multicolumn{1}{c|}{200}&14.54&\textbf{9.34}&\multicolumn{1}{c|}{-5.20}&8.89&\textbf{9.49}&\multicolumn{1}{c|}{0.60}&6.97&10.87&3.89&45.19&\textbf{51.96}&\multicolumn{1}{c|}{6.77}&39.66&\textbf{34.38}&\multicolumn{1}{c|}{-5.28}&24.77&46.52&21.76\\
\hline
\multicolumn{1}{c|}{\multirow{3}{*}{FT+L}}&\multicolumn{1}{c|}{1}&-2.30&\textbf{1.27}&\multicolumn{1}{c|}{{\ul 3.57}}&-0.52&\textbf{0.07}&\multicolumn{1}{c|}{{\ul 0.59}}&1.86&\textbf{-0.66}&{\ul -2.52}&100.81&&\multicolumn{1}{c|}{}&27.81&&\multicolumn{1}{c|}{}&6.59&24.30&17.71\\
\multicolumn{1}{c|}{}&\multicolumn{1}{c|}{50}&-3.43&-2.87&\multicolumn{1}{c|}{0.56}&-2.71&-3.07&\multicolumn{1}{c|}{-0.36}&-0.70&-2.35&-1.65&18.92&15.75&\multicolumn{1}{c|}{-3.17}&19.79&14.56&\multicolumn{1}{c|}{-5.24}&7.19&22.21&15.01\\
\multicolumn{1}{c|}{}&\multicolumn{1}{c|}{200}&-2.59&-3.44&\multicolumn{1}{c|}{-0.84}&-3.60&-3.81&\multicolumn{1}{c|}{-0.21}&-0.92&-2.99&-2.07&-2.45&-2.60&\multicolumn{1}{c|}{-0.15}&-0.74&-3.64&\multicolumn{1}{c|}{-2.91}&2.71&-1.96&-4.67\\
\hline
\multicolumn{1}{c|}{\multirow{3}{*}{MEND}}&\multicolumn{1}{c|}{1}&0.86&\textbf{0.72}&\multicolumn{1}{c|}{-0.14}&0.07&-0.69&\multicolumn{1}{c|}{-0.76}&1.66&\textbf{-0.07}&{\ul -1.73}&1.29&&\multicolumn{1}{c|}{}&-0.11&&\multicolumn{1}{c|}{}&1.51&0.29&{\ul -1.22}\\
\multicolumn{1}{c|}{}&\multicolumn{1}{c|}{\cancel{50}}&360.75&493.50&\multicolumn{1}{c|}{\cancel{132.75}}&427.41&450.42&\multicolumn{1}{c|}{\cancel{23.01}}&137.39&455.15&\cancel{317.75}&89.14&76.42&\multicolumn{1}{c|}{\cancel{-12.72}}&71.07&70.72&\multicolumn{1}{c|}{\cancel{-0.36}}&45.04&75.22&\cancel{30.19}\\
\multicolumn{1}{c|}{}&\multicolumn{1}{c|}{\cancel{200}}&361.28&401.57&\multicolumn{1}{c|}{\cancel{40.29}}&398.28&248.82&\multicolumn{1}{c|}{\cancel{-149.46}}&170.13&459.61&\cancel{289.48}&428.39&390.63&\multicolumn{1}{c|}{\cancel{-37.76}}&340.96&280.78&\multicolumn{1}{c|}{\cancel{-60.17}}&150.13&442.50&\cancel{292.37}\\
\hline
\multicolumn{1}{c|}{\multirow{3}{*}{ROME}}&\multicolumn{1}{c|}{1}&-2.20&\textbf{1.05}&\multicolumn{1}{c|}{{\ul 3.24}}&-0.23&\textbf{}&\multicolumn{1}{c|}{-0.33}&4.69&\textbf{-0.96}&{\ul -5.65}&-1.19&&\multicolumn{1}{c|}{}&0.89&&\multicolumn{1}{c|}{}&6.33&\textbf{-0.06}&{\ul -6.39}\\
\multicolumn{1}{c|}{}&\multicolumn{1}{c|}{\cancel{50}}&99.09&81.91&\multicolumn{1}{c|}{\cancel{-17.18}}&83.84&77.07&\multicolumn{1}{c|}{\cancel{-6.77}}&64.81&90.04&\cancel{25.22}&921.70&980.84&\multicolumn{1}{c|}{\cancel{59.14}}&1016.98&1001.62&\multicolumn{1}{c|}{\cancel{-15.36}}&665.52&994.84&\cancel{329.32}\\
\multicolumn{1}{c|}{}&\multicolumn{1}{c|}{\cancel{200}}&226.50&204.29&\multicolumn{1}{c|}{\cancel{-22.21}}&201.34&197.18&\multicolumn{1}{c|}{\cancel{-4.16}}&229.24&230.52&\cancel{1.28}&386.17&359.52&\multicolumn{1}{c|}{\cancel{-26.65}}&461.55&346.48&\multicolumn{1}{c|}{\cancel{-115.08}}&244.37&415.69&\cancel{171.33}\\
\hline
\multicolumn{1}{c|}{\multirow{3}{*}{MEMIT}}&\multicolumn{1}{c|}{1}&0.62&\textbf{0.48}&\multicolumn{1}{c|}{-0.14}&0.32&\textbf{0.61}&\multicolumn{1}{c|}{{\ul 0.28}}&-4.10&0.34&4.44&-2.73&&\multicolumn{1}{c|}{}&-0.17&&\multicolumn{1}{c|}{}&2.31&\textbf{-0.32}&{\ul -2.63}\\
\multicolumn{1}{c|}{}&\multicolumn{1}{c|}{50}&-0.65&-0.19&\multicolumn{1}{c|}{{\ul 0.46}}&-0.75&-0.34&\multicolumn{1}{c|}{{\ul 0.41}}&2.70&\textbf{-0.79}&{\ul -3.49}&-0.38&\textbf{0.85}&\multicolumn{1}{c|}{{\ul 1.23}}&-0.20&-0.52&\multicolumn{1}{c|}{-0.32}&3.26&\textbf{-1.06}&{\ul -4.31}\\
\multicolumn{1}{c|}{}&\multicolumn{1}{c|}{200}&-0.66&\textbf{1.18}&\multicolumn{1}{c|}{{\ul 1.83}}&0.34&\textbf{0.31}&\multicolumn{1}{c|}{-0.03}&2.61&\textbf{-0.80}&{\ul -3.41}&1.08&\textbf{1.54}&\multicolumn{1}{c|}{{\ul 0.46}}&1.42&\textbf{0.45}&\multicolumn{1}{c|}{-0.97}&4.05&0.86&{\ul -3.19}\\
\hline
\multicolumn{1}{c|}{\multirow{3}{*}{SIR\_top5}}&\multicolumn{1}{c|}{1}&-0.067&\textbf{2.863}&\multicolumn{1}{c|}{{\ul2.93}}&-0.499&-0.94&\multicolumn{1}{c|}{-0.441}&3.261&\textbf{-1.399}&{\ul-4.66}&-3.631&&\multicolumn{1}{c|}{}&-0.213&&\multicolumn{1}{c|}{}&2.617&\textbf{-0.35}&{\ul-2.967}\\
\multicolumn{1}{c|}{}&\multicolumn{1}{c|}{50}&-0.322&-0.916&\multicolumn{1}{c|}{-0.594}&-0.727&-0.073&\multicolumn{1}{c|}{{\ul0.654}}&2.511&\textbf{-1.418}&{\ul-3.929}&-0.634&\textbf{0.682}&\multicolumn{1}{c|}{{\ul1.316}}&-0.239&-0.699&\multicolumn{1}{c|}{-0.460}&2.064&\textbf{-1.098}&{\ul-3.162}\\
\multicolumn{1}{c|}{}&\multicolumn{1}{c|}{200}&-0.804&\textbf{0.979}&\multicolumn{1}{c|}{{\ul1.783}}&0.331&\textbf{0.544}&\multicolumn{1}{c|}{{\ul0.213}}&2.219&\textbf{-0.815}&{\ul-3.034}&0.318&\textbf{1.417}&\multicolumn{1}{c|}{{\ul1.099}}&-0.329&-0.527&\multicolumn{1}{c|}{-0.198}&1.241&\textbf{-0.476}&{\ul-1.717}\\\hline
\multicolumn{1}{c|}{\multirow{3}{*}{SIR\_top10}}&\multicolumn{1}{c|}{1}&-0.106&\textbf{2.148}&\multicolumn{1}{c|}{{\ul2.254}}&-0.706&-0.933&\multicolumn{1}{c|}{-0.227}&2.544&\textbf{-1.394}&{\ul-3.938}&-1.365&&\multicolumn{1}{c|}{}&-0.117&&\multicolumn{1}{c|}{}&2.048&\textbf{-0.384}&{\ul-2.432}\\
\multicolumn{1}{c|}{}&\multicolumn{1}{c|}{50}&-0.406&\textbf{0.578}&\multicolumn{1}{c|}{{\ul0.984}}&-0.939&-0.065&\multicolumn{1}{c|}{{\ul0.874}}&1.29&\textbf{-1.552}&{\ul-2.842}&-0.57&\textbf{0.565}&\multicolumn{1}{c|}{{\ul1.135}}&-0.281&-1.216&\multicolumn{1}{c|}{-0.935}&2.489&\textbf{-0.985}&{\ul-3.474}\\
\multicolumn{1}{c|}{}&\multicolumn{1}{c|}{200}&-0.838&\textbf{0.947}&\multicolumn{1}{c|}{{\ul1.785}}&0.339&\textbf{0.481}&\multicolumn{1}{c|}{{\ul0.142}}&1.772&\textbf{-0.728}&{\ul-2.5}&0.234&\textbf{1.421}&\multicolumn{1}{c|}{{\ul1.187}}&-0.054&-0.31&\multicolumn{1}{c|}{-0.256}&1.528&\textbf{-0.575}&{\ul-2.103}\\\hline
\end{tabular}
}
\vspace{-2mm}
\caption{Comparative analysis of perplexity changes. 
The first row represents the density of different editing targets (where BFS produces denser editings compared to Random). 
The second row reflects the impact observed at varying distances from the editing target, with ``inf'' denoting nodes that are not connected to the editing target. 
Note that the Vanilla evaluation is based on the original KG, whereas the GIE method is based on the similarity KG, so the distances (hops) are calculated within each respective graph.
The editing methods are listed in the leftmost column, and the adjacent column specifies the number of edits applied. 
The slashed values indicate instances where the method is incapable of handling the specified number of edits. 
Underlined values indicate that the ripple effect in the hidden space is more pronounced compared to the other two variants.
Bolded values highlight GIE detect a more heavier ripple effect than vanilla method.
}
\label{tab:simple_experiment}
\vspace{-5mm}
\end{table*}

%% file: latexs/06Experiment.4.tex
\input{tables/error_approximation_table}

\begin{figure*}[t]
    \centering
    \includegraphics[width=0.245\linewidth]{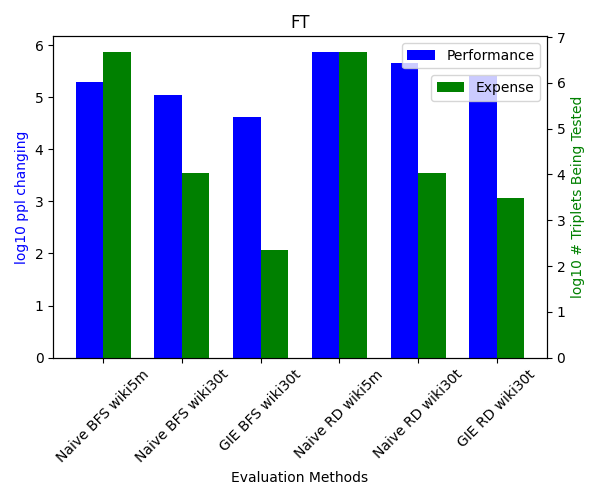}\hfill
    \includegraphics[width=0.245\linewidth]{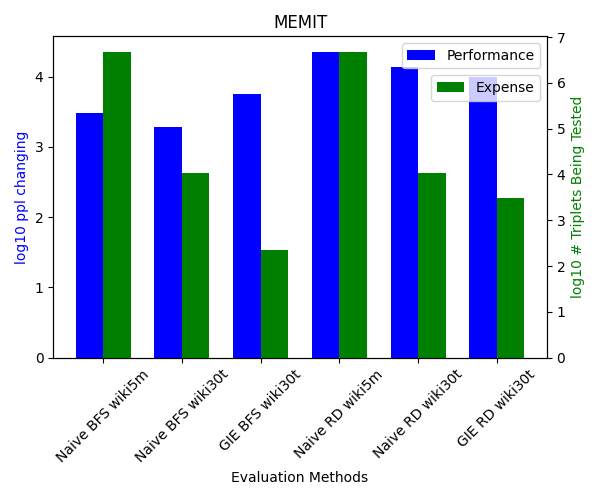}\hfill
    \includegraphics[width=0.245\linewidth]{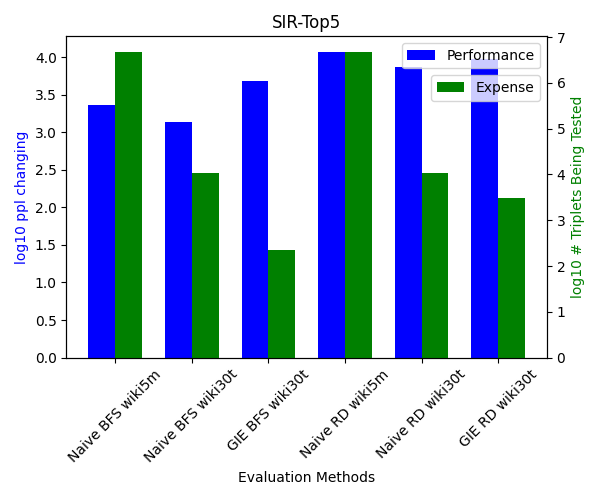}\hfill
    \includegraphics[width=0.245\linewidth]{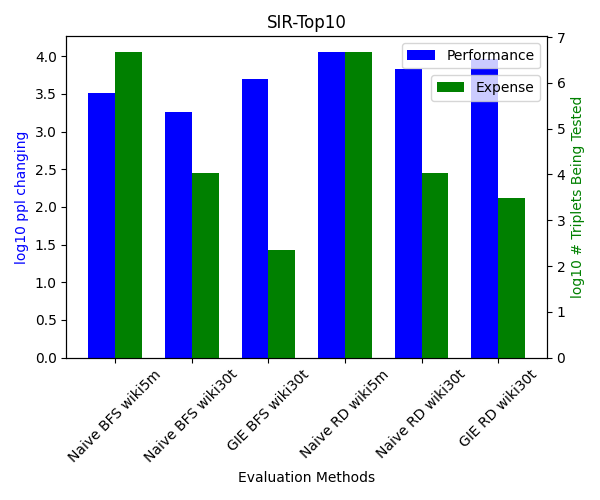}\hfill
    \caption{The \textbf{blue bar} represents the total change in perplexity across all baselines and evaluation methods. In contrast, the \textbf{green bar} reflects the evaluation cost for all baselines across these methods.}
    \label{fig:sum_ppl_chng}
\end{figure*}

\subsection{Overall Ripple Effects Evaluation}

\textbf{Ripple Effect Differs on Both Different Edit Quantities and Distributions.}  
As shown in Tab.~\ref{tab:simple_experiment}, the evaluation results indicate that model performance is influenced by both the quantity and the distribution of the edits. The intensity of the ripple effect escalates with an increasing number of edits; under identical edit quantities, the ripple effect is generally more pronounced in breadth-first search (BFS) distributed edits than randomly distributed edits.

\textbf{Excessive Edits Lead to Model Deterioration.}  
The performance of ROME and MEND significantly deteriorates when the number of edits exceeds 50. Although FT+L appears stable in Tab.~\ref{tab:simple_experiment}, it is not a practical approach as its updating mechanism restricts weight adjustments, thereby hindering the efficient update of parameters and the generation of meaningful sentences, as evidenced in Tab.~\ref{tab:case2}.

\textbf{GIE Detects Preciser Ripple Effects on Most Baselines than Vanilla.}
From Tab.\ref{tab:simple_experiment}, it is evident that proximity within the original KG used by Vanilla does not always result in a more significant ripple effect when compared to the similarity graph used by GIE.
This challenges the assumption that ``closer nodes are necessarily more affected by editing''~\cite{cohen2023evaluating}.
We conclude that no consistent correlation exists between distance on the original KG and reduced model performance.
Both nearby and distant triplets experience substantial ppl changes following model edits.

As demonstrated by the bolded values in Tab.~\ref{tab:simple_experiment}, triplets closer to the edit target in the similarity graph exhibit more significant increases in perplexity compared to the original KG evaluated by Vanilla, whereas nodes with no direct connections show a minor increase in perplexity.
This demonstrates the higher sensitivity of GIE in detecting ripple effects.
The underlined values in Tab.~\ref{tab:simple_experiment} highlight the differences in perplexity changes between GIE, which uses the similarity graph, and Vanilla, which uses the original KG. 
For example, in the BFS method under a single edit, the distinction between the similarity graph in GIE and the original KG in Vanilla is particularly notable. This shows that GIE captures the ripple effects more effectively by considering the semantic relationships between triplets.
In contrast, the ripple effects observed in Vanilla, which relies on direct connections in the original KG, are often less pronounced or damaging, as seen in subsequent entries of the same row.

\textbf{GIE Detects 90\% of the Overall Ripple Effect with Only 10\% of the Computational Cost Compare to Vanilla}
As illustrated in Fig.~\ref{fig:sum_ppl_chng}, we analyzed the overall ppl changes caused by four baseline methods on the entire KG.
Compared to the average ppl change calculated in Tab.~\ref{tab:simple_experiment}, the Vanilla method captured a ripple effect closer to the total observed ripple effect.
However, GIE was able to detect 90\% of the ripple effect with only 10\% of the computational cost required by the Vanilla method.
Moreover, with advanced model editing techniques such as MEMIT and SIR, the overall ppl may be lower than the local ppl.
This is because most of the knowledge is retained by the models, reducing ppl.

\subsection{In-depth Comparison Between Vanilla and GIE Evaluation}

\begin{figure}[t]
\centering
\includegraphics[width=\columnwidth]{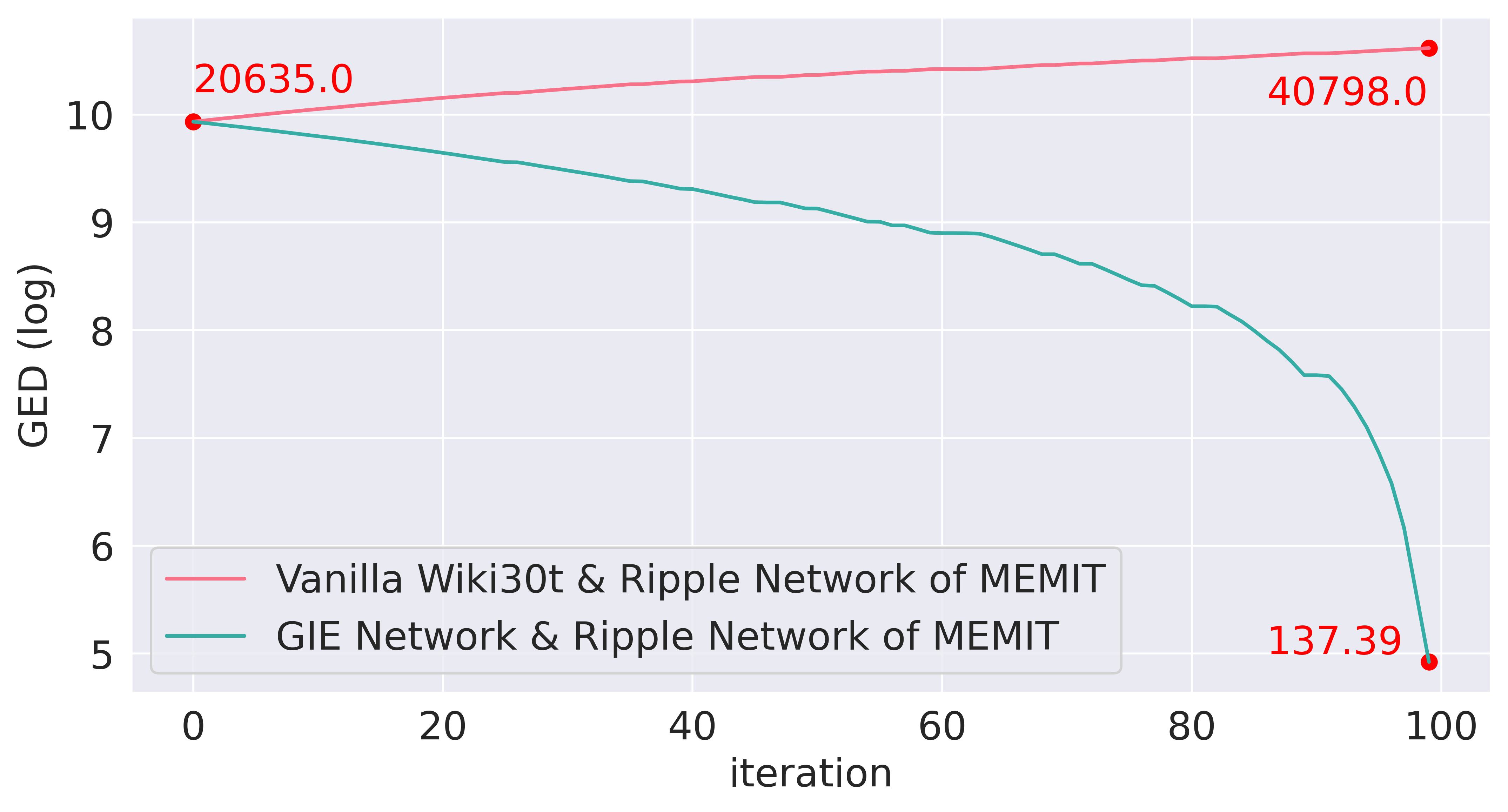}
\caption{The GED's change, with the x-axis representing the iterations of building the Ripple Network of MEMIT.
The higher the score is, the more structural difference the two graphs have.
}
\label{fig.2}
\vspace{-5mm}
\end{figure}
\begin{figure}[t]
\begin{center}
\includegraphics[width=\columnwidth]{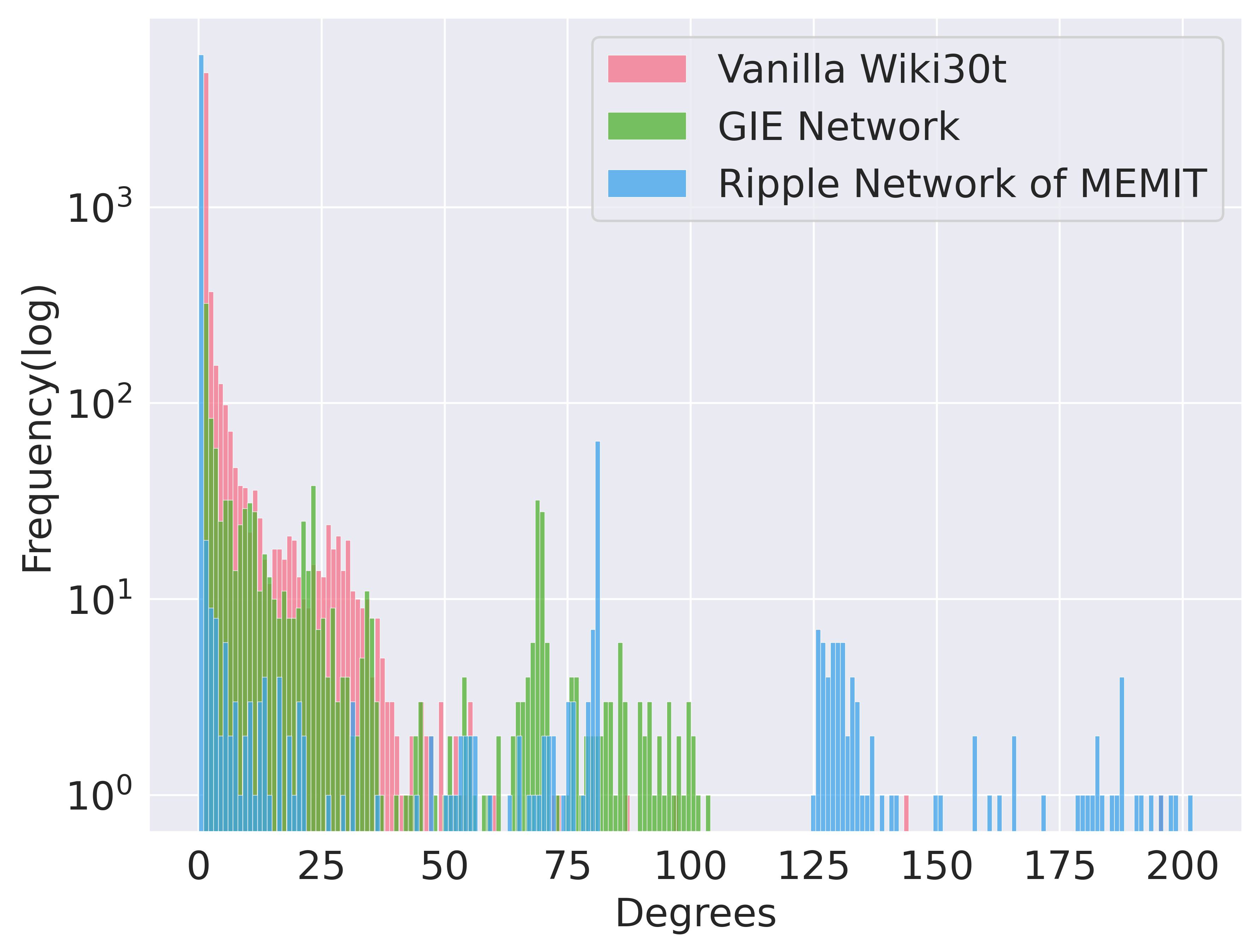}
\end{center}
\vspace{-5mm}
\caption{the frequency of node degrees within the vanilla KG, GIE network, and Ripple Network of MEMIT.}
\label{fig.3}
\end{figure}

We embed the overall ripple effect to the original KG and compare them to the original KG and the GIE similarity graph~(we call it \textbf{GIE network} here). 
This allows us to \textbf{visualize the differences} between the Vanilla and GIE evaluation methods and the actual ripple effect.

To explore this, we employed the state-of-the-art model editing method MEMIT to construct a ripple network, referred to as the \textbf{Ripple Network of MEMIT}. 
This network is built iteratively by editing a specific triplet in the KG and connecting the most affected entities to the edited entity, thereby visualizing the scope of the edits' impact through the network.
We edited 100 triples in sequence to ensure that the number of edges in the Ripple Network is as similar as possible to that of the original KG, allowing for a comparable scale across the three networks.

We utilized the Graph Edit Distance (GED) to quantify these networks' differences further. This metric evaluates the impact of changes on the structural integrity and information consistency of the knowledge graph. We employed a simplified version of GED, calculated using the L1 norm:
\[
\text{GED} = \log\left( \left| \mathbf{G}_{\text{adj}} - \mathbf{G'}_{\text{adj}} \right| \right),
\]
where $\mathbf{G}_{\text{adj}}$ and $\mathbf{G'}_{\text{adj}}$ represent the adjacency matrices of the two graphs.

Figure~\ref{fig.2} illustrates how the differences between the networks change as the number of knowledge edits increases.
The higher initial GED between the Ripple Network and the other two networks is because, at iteration 0, the Ripple Network contains only nodes and no edges.
As the iterations progress, the GED between the Ripple Network and the GIE Network decreases, indicating that the structure of the GIE Network increasingly resembles that of the Ripple Network. 
This suggests that the GIE evaluation method effectively captures the most impacted aspects of the ripple effect.
In contrast, the GED between the Ripple Network and the original KG continues to increase, indicating that the original KG contains many irrelevant connections and is less suitable for detecting ripple effects.

Figure~\ref{fig.3} shows the degree distribution of nodes across the three networks.
The original KG exhibits a steep drop in the frequency of high-degree nodes, a common characteristic in real-world networks.
In contrast, the Ripple Network of MEMIT has a more uniform distribution of node degrees, suggesting a more balanced connectivity distribution. 
Interestingly, the structure of the GIE Network more closely resembles that of the original KG than the Ripple Network, indicating that while the GIE method captures some aspects of the ripple effect, it does not precisely reflect the full extent of its impact. This suggests that there is room for further improvement in the GIE approach.

\begin{figure}[t]
\centering
\includegraphics[width=\columnwidth]{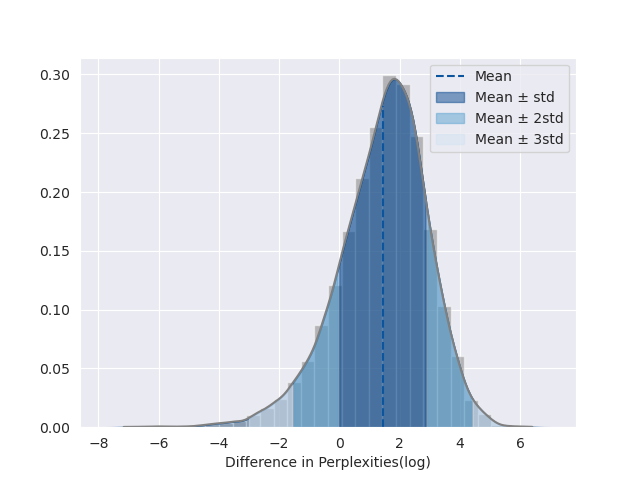}
\caption{
The frequency distribution of perplexity changes after model editing.
}
\label{fig:metric_dis}
\end{figure}
\begin{figure}[t]
\begin{center}
\includegraphics[width=\columnwidth]{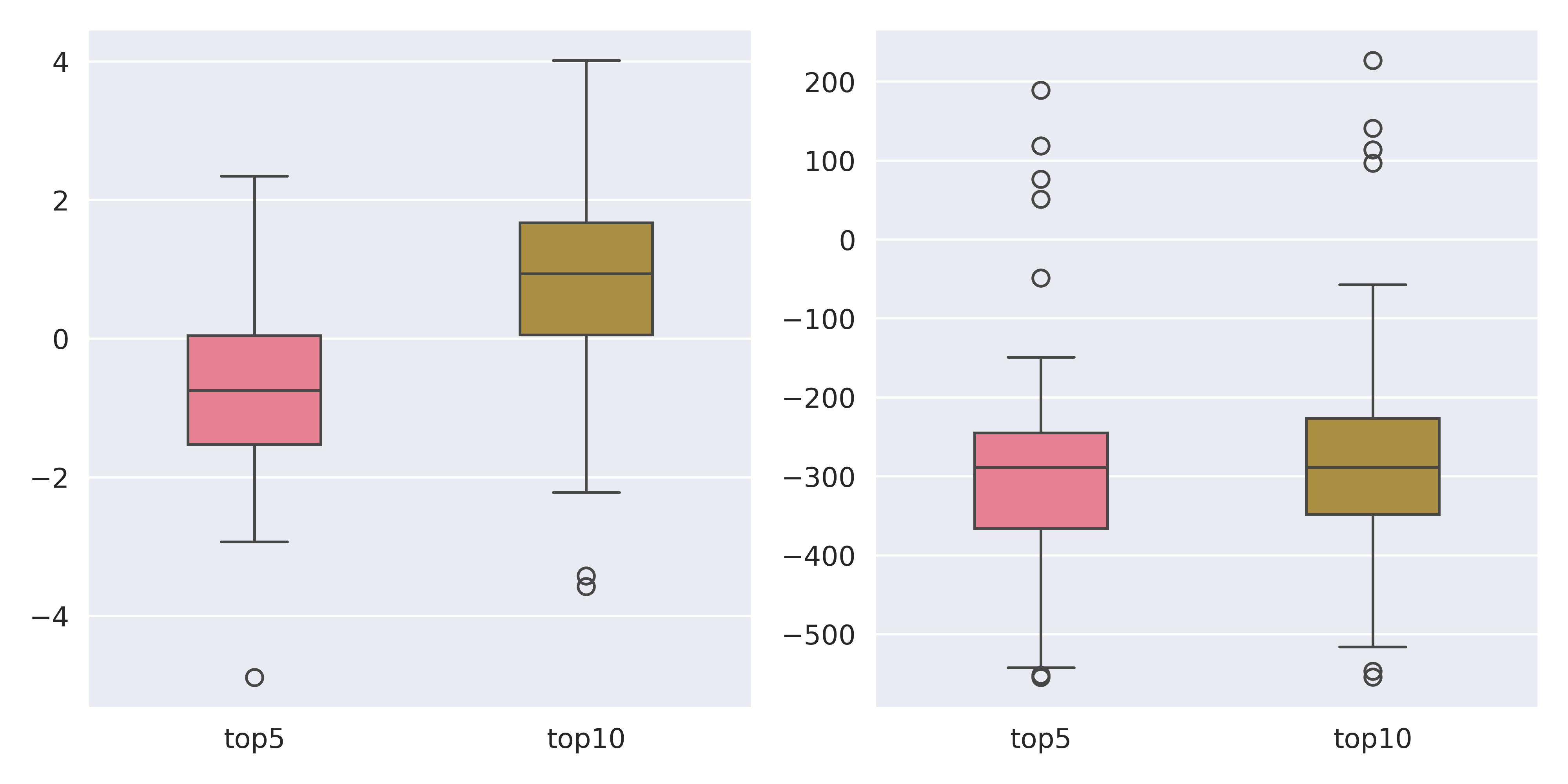}
\end{center}
\caption{Average changing in perplexity attributed to SIR.
The left panel shows the overall perplexity change, while the right panel shows the perplexity change for the triplets most similar to the edit targets.}
\label{fig.4}
\vspace{-7mm}
\end{figure}

\input{tables/case_study_table}

\subsection{In-depth Analysis of SIR Based on Perplexity Changing}

Fig.~\ref{fig:metric_dis} presents the frequency distribution of perplexity changes before and after typical model edits.
The figure suggests that these changes in the evaluation metric approximately follow a normal distribution.
Hence, a triplet with significant perplexity change can be defined as one where the change exceeds a certain threshold: $ \delta > \mu + 2\sigma $. 
Here, $\delta$ denotes the change in perplexity before and after editing, $\mu$ is the mean, and $\sigma$ is the standard deviation.
So, we find that \textbf{only a few triplets exhibit significant changes} in perplexity during one single model editing. 

Therefore, SIR mitigates the ripple effect by selectively re-editing a small subset of triplets.
We assess the efficacy of the SIR method by comparing the re-editing of different numbers of the top-K triplets that are most similar to the edit targets.
As illustrated in Fig.~\ref{fig.4}, \textbf{re-editing the top-5 triplets substantially reduces overall perplexity}, with a particularly marked improvement for these specific triplets. 
However, extending the re-edits to the top 10 triplets slightly increases overall perplexity due to the complexities introduced by numerous edits.

\subsection{Case Study}

In Tab.~\ref{tab:case}, we investigate the text changes generated by GPT2-XL in response to an edit request, focusing on the sentences among the top 10 triplets with embeddings most similar to the edit target.
Before editing, the model generates accurate and coherent content; however, after editing, a subset of the outputs, identified as the triplets that have the most similar embedding to the edit target by GIE, contain incorrect or nonsensical samples.
Employing SIR enables the model to generate accurate results once again.
Nevertheless, since the third fact was not among the top 5 triplets with embeddings most similar to the edit target, it was not re-edited in SIR-top5, causing the model to maintain the same outputs for that fact.
Tab.~\ref{tab:case2} illustrates that when handling multiple edits, FT, FT+L, MEND, and ROME cause severe model crashes.
The model generates repetitive word patterns and fails to produce coherent sentences, rendering quantitative assessment impractical, leading us to strike out the result in Tab.~\ref{tab:experiment}.

%% file: tables/error_approximation_table.tex

%% file: tables/case_study_table.tex
\begin{table*}[t]

    \centering
\noindent
\begin{minipage}{\columnwidth}
    \centering
    \tiny
    \begin{tabular}{>{\raggedright\arraybackslash}p{7cm}}
        \toprule
        \textbf{Samples Generated by GPT2-XL}\\
        \hline
        \textbf{Edit request} \\
        (Ethiopia, member of,\textcolor{purple}{ United Nations)} $\to$ 
        (Ethiopia, member of, \textcolor{blue}{European Union)}
        \\ \hdashline
        \textbf{Pre-Edit} \\
        The given name of Elizabeth Christ is a common name in the United States.
        Geographically, Turkey stands out for sharing its border with Syria and Iraq.
        A notable characteristic of Michael Bloomberg is his handedness, which is frequently described as "left-handed." 
        \\
        \hline
        \textbf{Post-Edit} 
        \\ \hdashline
        \textbf{MEMIT}\\
        The given name of Elizabeth Christ is \textcolor{red}{Elizabeth Stedman}, she is the founder and editor of Christ and Pop Culture. \xmark \\
        Geographically, Turkey stands out for sharing its border with \textcolor{red}{Turkey}, but is not part of the EU. \xmark \\
        A notable characteristic of Michael Bloomberg is his handedness, which is of his left \textcolor{red}{eye}. \xmark \\
        \\ \hdashline
        \textbf{ICE} 
        The given name of Elizabeth Christ is  a common English first name for a woman, and it is used in \textcolor{red}{Ethiopia} as a first name, and in the UK as a surname. \xmark \\ 
        Geographically, Turkey stands out for sharing its border with  \textcolor{red}{Ethiopia} and has been a key transit point for African migrants.\xmark \\ 
        A notable characteristic of Michael Bloomberg is his handedness, which is vernacular for "lefty", but it is not his only one. \cmark \\ 
        
        \\ \hdashline
        \textbf{SIR(top5)} \\
        The given name of Elizabeth Christ is \textcolor{green}{Elizabeth}. \cmark \\ 
        Geographically, Turkey stands out for sharing its border with \textcolor{green}{Syria}. \cmark \\
        A notable characteristic of Michael Bloomberg is his handedness, which is of his left \textcolor{red}{eye}. (not re-edited) \xmark \\
        \\ \hdashline
        \textbf{SIR(top10)}\\
        The given name of Elizabeth Christ is \textcolor{green}{Elizabeth}. \cmark \\ 
        Geographically, Turkey stands out for sharing its border with \textcolor{green}{Syria}. \cmark \\
        A notable characteristic of Michael Bloomberg is his handedness, which is \textcolor{green}{left-handed}. \cmark \\
    \bottomrule
    \end{tabular}
    \caption{\small{Case Study of text Generated by GPT2-XL with and without SIR implementation.}}
    \label{tab:case}
\end{minipage}
\begin{minipage}{\columnwidth}
    \centering
    \tiny
    \begin{tabular}{>{\raggedright\arraybackslash}p{7cm}}

    \toprule
    \textbf{Samples Generated by Crashed Model} 
    \\ \hline
    \textbf{FT(50 edits)}\\
    The given name of Elizabeth Christ is \textcolor{red}{name of Chrispher}
    \textcolor{red}{Columbus Christ is a common European name.}
    Geographically, Turkey stands out for sharing its border with \textcolor{red}{ulov 150101 Crimean Tatar Kazakh Kazakhstan} \textcolor{red}{Kosovo Kyrgyzstan Lao People's Democratic Repub$\dots$}
    A notable characteristic of Michael Bloomberg is his handedness, which is \textcolor{red}{ia of the city.}
    \\ \hdashline
    \textbf{FT+L(10 edits)}\\
    The given name of Elizabeth Christ is \textcolor{red}{is is is is is is is is} \textcolor{red}{ is is is is is $\dots$}
    Geographically, Turkey stands out for sharing its border with  \textcolor{red}{(((((((((( I I IIIIII$\dots$}
    A notable characteristic of Michael Bloomberg is his handedness, which is \textcolor{red}{urch ourchurchurchurchurchur$\dots$}
    \\ \hdashline
    \textbf{MEND(50 edits)}\\
    The given name of Elizabeth Christ is \textcolor{red}{the" for@","@ the-}
    \textcolor{red}{" for the $\dots$}
    Geographically, Turkey stands out for sharing its border with \textcolor{red}{"))")"))"))","@",""," and and $\dots$}
    A notable characteristic of Michael Bloomberg is his handedness, which is \textcolor{red}{nd for") on"))@@"@ the"))" the--$\dots$"}
    \\ \hdashline
    \textbf{ROME(50 edits)}\\
    The given name of Elizabeth Christ is \textcolor{red}{Winia Ss- stick }
    \textcolor{red}{event set S Beef Beeflde Avg$\dots$}
    Geographically, Turkey stands out for sharing its border with\textcolor{red}{Noinia the remotely Avg Medalinia Fó4 crank Tat $\dots$}
    A notable characteristic of Michael Bloomberg is his handedness, which is \textcolor{red}{theó Avg Avg Avg Avg Avg Avg Avg $\dots$}
    \\
    \bottomrule
    \toprule
    \textbf{Samples Generated by SIR-edited Moddel}
    \\ \hdashline
    \textbf{SIR(top5, 200 Edits)}
    The given name of Elizabeth Christ is Elizabeth  Ann Christ.
    
    Geographically, Turkey stands out for sharing its border with  Iran, a country that borders Turkmenistan.

    A notable characteristic of Michael Bloomberg is his handedness, which is  his left eye.
    \\ \hdashline
    \textbf{SIR(top10, 200 Edits)} \\
    The given name of Elizabeth Christ is Elizabeth Ann Christ.
    
    Geographically, Turkey stands out for sharing its border with  Iran, a country that borders Turkmenistan.

    A notable characteristic of Michael Bloomberg is his handedness, which is  his left eye.
    \\
    \bottomrule
    \end{tabular}
    \caption{\small{Cases for different editing methods dealing with multiple edits.}}
    \label{tab:case2}
\end{minipage}
\end{table*}


%% file: latexs/07Conclusion.2.tex
In conclusion, this paper has made significant strides in understanding and mitigating the ripple effect in the hidden space, a complex and challenging issue in editing LLMs.
We have proposed an innovative evaluation methodology, Graphical Impact Evaluation~(GIE), which effectively identifies the ripple effect in the hidden space during model editing.
Furthermore, we have developed a novel model editing method, the Selective Impact Re-Editing Approach~(SIR), which leverages the design of GIE to mitigate the ripple effect in the hidden space.
Our comprehensive evaluations and comparative experiments have demonstrated the effectiveness of both GIE and SIR.
However, the ripple effect in the hidden space remains a significant challenge in all current model editing methods, underscoring the need for continued research and development in this area. 

%% file: latexs/Limitation.tex
\textbf{Efficiency}
Our approach involves editing and evaluating based on a KG. Owing to the large scale of KG, this process is both time-intensive and demands substantial computational resources. 

\textbf{Dependence on KGs}
Our methodology relies on KGs. However, ensuring the quality of these graphs proves to be a complex task. Evaluating KGs in practical scenarios presents many challenges.

\textbf{Model Selection}
Given the constraints of computational resources, our analysis has been limited to GPT2-XL. However, the effectiveness of our method for models of varying sizes and architectures needs further investigation.

%% file: latexs/09Appendix.1.tex
\subsection{Detail Experiments}
We put the detail experiments with detail setting and more baselines.
These baselines include several approaches that rely on external storage mechanisms, while do not change the parameter of the original models. 
The detailed experimental results are presented in Tab.~\ref{tab:experiment}
And the description of these additional baselines are as follow:
\textbf{Semi-Parametric Editing with a Retrieval-Augmented Counterfactual Model~(SERAC)}~\cite{cohen2023evaluating} stores user-provided edits in an explicit memory and uses a scope classifier and counterfactual model to modulate the base model's predictions without modifying its parameters.
\textbf{Edit models by REading Notes(EREN}~\cite{cohen2023evaluating} stores all edits in a notebook memory and retrieves relevant notes to modify the behavior of large language models based on context, without altering model parameters.
\textbf{In-context Editing~(ICE)}~\cite{cohen2023evaluating} prepend the following prefix to the input prompt: ``Imagine that <$O^*$> would have been <$P_r$>''.

\input{tables/table}

\subsection{Prompt}
\begin{table}[!ht]
    \centering
    \small

    \begin{tabular}{>{\raggedright\arraybackslash}p{0.9\columnwidth}}
    \hline
    \textbf{Prompt used in dataset construction} \\
    \hline
    \textbf{Prompt}\\
    In this case, I will provide a triplet (s, p, o), and I need you to design 3-5 prompts based on this triplet. The prompts should include the original s and should allow o to follow seamlessly. For example, if I give the triplet \{`s': `White House', `p': `architectural style', `o': `Neoclassical architecture'\}, your answer should be in JSON format like \{`s': `White House', `p': `architectural style', `o': `Neoclassical architecture', 'prompt': [`White House is designed in the architectural style of ', `The White House showcases the distinctive architectural style of ', `When discussing the architectural style of the White House, one immediately thinks of']\}. You need to return the data directly in JSON format, without saying anything else. This time, the triplet I provide is \{`s':`',`p':`',`o':`'\}.\\
    \hdashline
    \textbf{Example Triplet}\\
    \{\\
"s": "Washington, D.C.",\\
\space"p": "shares border with",\\
\space"o": "Virginia"\\
\}\\
    \hdashline
    \textbf{Response}\\
\{\\
"s": "Washington, D.C.",\\
"p": "shares border with",\\
"o": "Virginia",\\
"prompt": [
"Washington, D.C. is known for sharing its border with ",
"A key geographical feature of Washington, D.C. is its border with ",
"Discussing the borders of Washington, D.C., one commonly mentions its adjacency to ",
"An important aspect of Washington, D.C.'s location is its shared border with ",
"In the context of regional boundaries, Washington, D.C. is notably adjacent to "\
]
\}\\

    \hline
    \end{tabular}
    \caption{Example of prompt generation based on a given triplet for dataset construction.}
    \label{tab:prompt}
\end{table}

In constructing our dataset, we utilize GPT4 to generate prompts that integrate specific subjects with their corresponding predicates. As illustrated in Tab.~\ref{tab:prompt}, this method ensures the quality and fluency of our data.

We also utilize GPT4 to generate ICE prefix prompts. Tab.~\ref{tab:ICE} shows an example.

\begin{table}[!ht]
    \centering
    \small
    \begin{tabular}{>{\raggedright\arraybackslash}p{0.9\columnwidth}}
    \hline
    \textbf{Prompt used for ICE} \\
    \hline
    \textbf{Prompt}\\
    In this case, I will give you a json, please help me to output it in subjunctive mood. For example: given  \{"prompt": "\{\} is a relative of ", "subject": "Donald Trump", "target":  "Glenn D'Hollander"\}. You need to output "Imagine that Glenn D'Hollander would have been a relative of Donald Trump." This time, the json I provide is \{"prompt": "", "subject": "", "target": \} .\\
    \hdashline
    \textbf{Example JSON}\\
    \{\\
"prompt": "\{\} held the position of ",\\
\space"subject": "Donald Trump",\\
\space"target": "president of the Constitutional Court of Spain"\\
\}\\
    \hdashline
    \textbf{Response}\\
Imagine that Donald Trump had held the position of president of the Constitutional Court of Spain.\\
    \hline
    \end{tabular}
    \caption{Example of prefix prompt generation for ICE.}

    \label{tab:ICE}
\end{table}

\subsection{Model Selection}
Due to the limitation of computation resources, we perform experiments on GPT2-XL~\cite{radford2019language}. GPT-2 XL is the 1.5B parameter version of GPT-2, a transformer-based language model created and released by OpenAI. The model is a pre-trained model on the English language using a causal language modeling (CLM) objective. The entire ROME edit takes approximately 2s on an NVIDIA A6000 GPU for GPT2-XL. MEMIT takes 3226.35 sec $\approx $ 0.90 hr for 10,000 updates on GPT-J.

\subsection{Implementation details}
\label{sec:implementation}
\textbf{FT / FT+L} For basic Fine-Tuning (FT), we follow~\cite{meng2022locating} re-implementation in their study, using Adam~\cite{muller2022instant} with early stopping to minimize $ -\log \mathbb{P}_{G'}[o^*| p]$, changing only $mlp_{\text{proj}}$ weights at selected layer 1. We use a learning rate of $5 \times 10^{-4}$ and early stop at a 0.03 loss.

For constrained fine-tuning (FT+L)~\cite{zhu2020modifying}, we add an $L_{\infty}$ norm constraint: $\left \| \theta_G - \theta_{G'} \right \|_{\infty} \leq \epsilon $. It is achieved in practice by clamping weights $\theta_{G'}$ to the $\theta_G \pm \epsilon$ range at each gradient step. We select layer 0 and $\epsilon = 5 \times 10^{-4}$. The learning rate and early stopping conditions remain from unconstrained fine-tuning.

\textbf{MEND}~\cite{mitchell2022fast}learn a rank-1 decomposition of the negative log-likelihood gradient of some subset of $\theta_G$. Hyperparameters are adopted from given default configurations.

\textbf{ROME}~\cite{meng2022locating}  conceptualizes the MLP module as a straightforward key-value store. We directly apply the code and MLP weight provided by the original paper and keep the default setting for hyperparameters. We perform the intervention at layer 18, and covariance statistics are collected using 100,000 Wikitext samples.

\textbf{MEMIT}~\cite{meng2023massediting} builds upon ROME to insert many memories by modifying the MLP weights of a range of critical layers. Using their code, we tested the MEMIT ability, and all hyperparameters followed the same default settings. For GPT2-XL, we choose layers $ = [3, 4, 5, 6, 7, 8]$. 

\textbf{SERAC}~\cite{madaan2022memory} introduces a semi-parametric model editing approach using a retrieval-augmented counterfactual model. We utilize the code provided by the original paper to replicate the model editing experiments. The SERAC editor is composed of three primary components: a memory to store edits, a scope classifier, and a counterfactual model to generate new predictions. For each test input, the scope classifier determines whether it falls within the scope of the stored edits. If the input is within scope, the counterfactual model generates an output based on the most relevant stored edit; otherwise, the base model's output is used. We retain the default hyperparameters provided by the authors, applying the scope classifier at layer 18, and use a set of 100,000 Wikitext samples for collecting covariance statistics.

\textbf{EREN}~\cite{chen2024robust} proposes a robust and scalable model editing method that complements large language models (LLMs) with a notebook storing all edits in natural text. For each input, the model determines whether it is relevant to any stored edit. If relevant, the retrieved edits are used as prompts to adjust the LLM's behavior accordingly; if irrelevant, the model relies on its parametric knowledge. The code provided by the authors was directly applied, with default hyperparameters. We employed a two-step inference process and a dual-encoder retrieval framework, retrieving the top-5 most relevant edits using the Contriever retriever. We retained the default setting for retrieval size and used FLAN-T5-XL as the base model for our experiments.

\textbf{ICE}~\cite{cohen2023evaluating} does not introduce changes to the model parameters, but prepend the following prefix to the input prompt: ``Imagine that <$O^*$> would have been <$P_r$>''. The prompts are generated using GPT4. See Tab.~\ref{tab:ICE} for an example. Due to input length constraints, we conducted experiments with edit amounts set to $[1,2,3,5,8,10]$.

\textbf{SIR} re-edit the topK outliers. We use MEMIT to perform re-editing. All hyperparameters follow the same default settings with MEMIT. We conducted experiments with K set to $[5, 10]$.

\subsection{Other metrics}
\label{sec:appendix2}
\begin{figure*}[!hb]
\begin{center}
\resizebox{0.95\textwidth}{!}{
\includegraphics{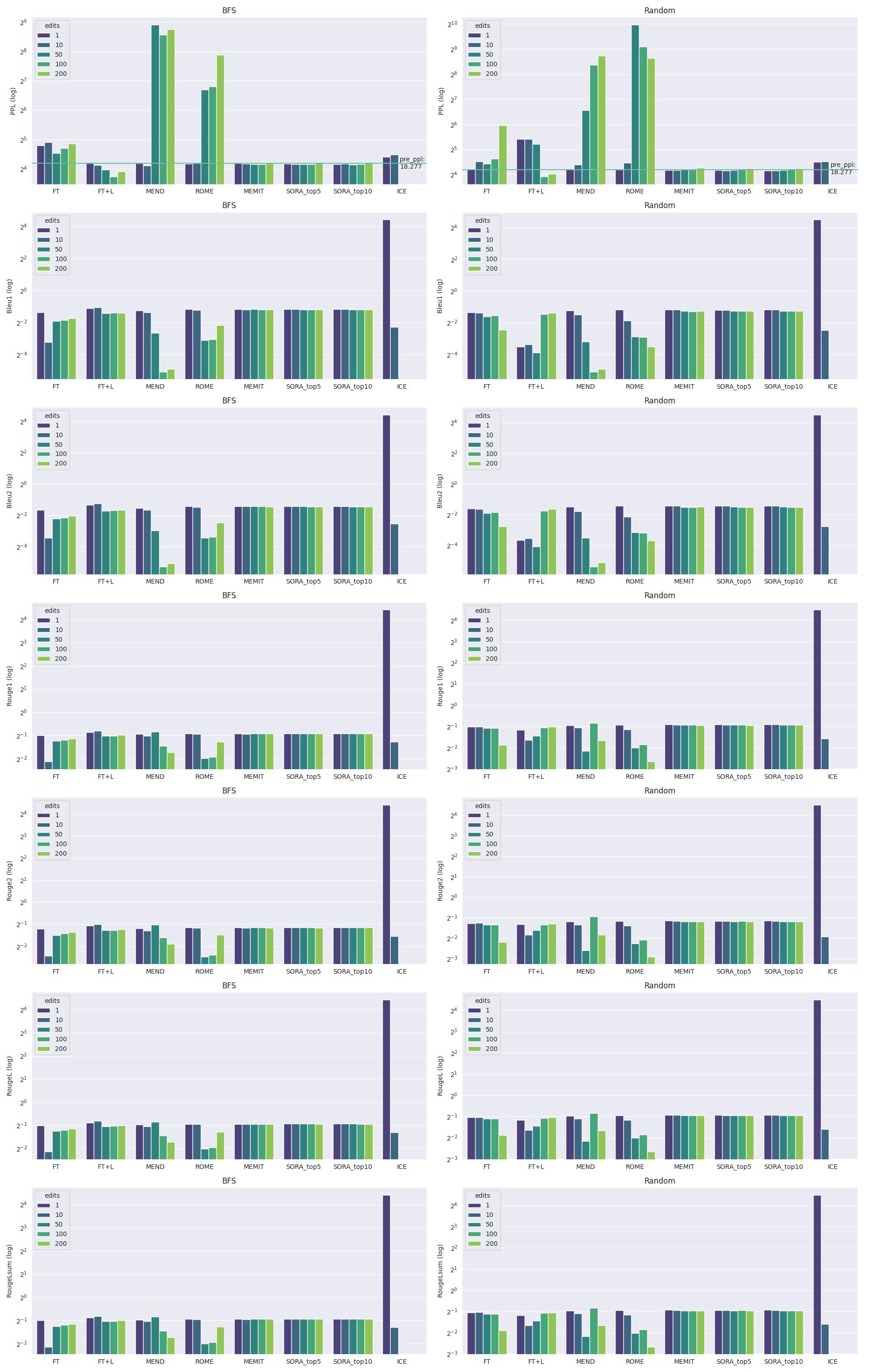}
}
\caption{
    Perplexity, Bleu and Rouge score.
}
\label{fig:appendix}
\end{center}
\vspace{-8mm}
\end{figure*}

We performed experiments utilizing alternative metrics. Fig.~\ref{fig:appendix} shows the detailed results. 
This set of bar graphs presents results across two different sampling strategies: Breadth-First Search (BFS) and Random sampling. 
Within each graph, model editing methods are compared. 
The bars are grouped by the number of edits, ranging from 1 to 200, with each group color-coded for clarity. The height of the bars corresponds to the metric's value on a logarithmic scale. 
In the PPL graphs, the horizontal line represents the average PPL of the dataset before model editing.
In the computation of BLEU and ROUGE metrics, the text generated by the post-edit model is employed as the Predictions. In contrast, the text generated by the original model serves as the Reference. It facilitates a comparative analysis of the discrepancies between the pre-edit and post-edit outputs.
After evaluating these metrics comparatively, we have selected PPL as the metric of choice for our experiment.

\subsection{License}
In the course of developing the methodologies and implementations detailed within this study, we have incorporated codes that are distributed under the terms of the MIT License~\footnote{https://github.com/kmeng01/memit}. It significantly bolstered our research, enabling us to focus on the novel contributions of our work without the necessity of developing foundational components from scratch. We extend our profound gratitude to the original authors for their invaluable contributions to the open-source community and affirm our commitment to adhering to the stipulations of the MIT License.

%% file: tables/table.tex
\begin{table*}[t]
\centering

\resizebox{\textwidth}{!}{
\begin{tabular}{cccccccccccccl|ccccccccl}
\hline
&&\multicolumn{12}{c|}{\textbf{BFS Sampling}}&\multicolumn{9}{c}{\textbf{Random Sampling}}\\\cline{3-23} 
&&\multicolumn{3}{c}{1}&\multicolumn{3}{c}{2}&\multicolumn{3}{c}{3}&\multicolumn{3}{c|}{inf}&\multicolumn{3}{c}{1}&\multicolumn{3}{c}{2}&\multicolumn{3}{c}{inf}\\\cline{3-23} 
Methods&\#Edition&Vanilla&GIE&Diff&Vanilla&GIE&Diff&Vanilla&GIE&Diff&Vanilla&GIE&\multicolumn{1}{c|}{Diff}&Vanilla&GIE&Diff&Vanilla&GIE&Diff&Vanilla&GIE&\multicolumn{1}{c}{Diff}\\
\hline
\multicolumn{1}{c|}{\multirow{5}{*}{FT}}&\multicolumn{1}{c|}{1}&5.77&\textbf{10.99}&\multicolumn{1}{c|}{{\ul 5.22}}&9.35&\textbf{8.97}&\multicolumn{1}{c|}{-0.38}&9.45&\textbf{8.89}&\multicolumn{1}{c|}{-0.56}&\multicolumn{1}{l}{10.54}&\multicolumn{1}{l}{8.94}&{\ul -1.60}&-7.09&&\multicolumn{1}{c|}{}&0.92&&\multicolumn{1}{c|}{}&\multicolumn{1}{l}{5.07}&\multicolumn{1}{l}{0.44}&{\ul -4.63}\\
\multicolumn{1}{c|}{}&\multicolumn{1}{c|}{10}&11.90&\textbf{10.69}&\multicolumn{1}{c|}{-1.20}&11.91&\textbf{10.65}&\multicolumn{1}{c|}{-1.26}&11.42&\textbf{12.23}&\multicolumn{1}{c|}{{\ul 0.82}}&5.42&12.86&7.44&4.27&\textbf{4.95}&\multicolumn{1}{c|}{{\ul 0.68}}&4.47&\textbf{4.55}&\multicolumn{1}{c|}{{\ul 0.08}}&14.95&4.23&{\ul -10.72}\\
\multicolumn{1}{c|}{}&\multicolumn{1}{c|}{50}&7.17&\textbf{4.65}&\multicolumn{1}{c|}{-2.52}&4.78&\textbf{3.89}&\multicolumn{1}{c|}{-0.89}&4.29&&\multicolumn{1}{c|}{}&3.73&5.23&1.50&3.21&\textbf{1.48}&\multicolumn{1}{c|}{-1.73}&1.92&\textbf{4.35}&\multicolumn{1}{c|}{2.43}&22.64&2.82&{\ul -19.82}\\
\multicolumn{1}{c|}{}&\multicolumn{1}{c|}{100}&12.80&\textbf{7.27}&\multicolumn{1}{c|}{-5.53}&6.89&\textbf{6.14}&\multicolumn{1}{c|}{-0.76}&6.72&&\multicolumn{1}{c|}{}&14.83&8.35&{\ul -6.48}&5.19&\textbf{5.15}&\multicolumn{1}{c|}{-0.04}&4.27&\textbf{1.77}&\multicolumn{1}{c|}{-2.50}&6.50&5.04&{\ul -1.47}\\
\multicolumn{1}{c|}{}&\multicolumn{1}{c|}{200}&14.54&\textbf{9.34}&\multicolumn{1}{c|}{-5.20}&8.89&\textbf{9.49}&\multicolumn{1}{c|}{0.60}&8.36&&\multicolumn{1}{c|}{}&6.97&10.87&3.89&45.19&\textbf{51.96}&\multicolumn{1}{c|}{6.77}&39.66&\textbf{34.38}&\multicolumn{1}{c|}{-5.28}&24.77&46.52&21.76\\
\hline
\multicolumn{1}{c|}{\multirow{5}{*}{FT+L}}&\multicolumn{1}{c|}{1}&-2.30&\textbf{1.27}&\multicolumn{1}{c|}{{\ul 3.57}}&-0.52&\textbf{0.07}&\multicolumn{1}{c|}{{\ul 0.59}}&1.17&\textbf{1.41}&\multicolumn{1}{c|}{{\ul 0.24}}&1.86&\textbf{-0.66}&{\ul -2.52}&100.81&&\multicolumn{1}{c|}{}&27.81&&\multicolumn{1}{c|}{}&6.59&24.30&17.71\\
\multicolumn{1}{c|}{}&\multicolumn{1}{c|}{10}&-3.15&-0.76&\multicolumn{1}{c|}{{\ul 2.39}}&-0.85&-0.14&\multicolumn{1}{c|}{{\ul 0.72}}&-0.20&-0.24&\multicolumn{1}{c|}{-0.04}&0.63&\textbf{-1.01}&{\ul -1.64}&33.22&\textbf{20.49}&\multicolumn{1}{c|}{-12.73}&24.11&\textbf{21.37}&\multicolumn{1}{c|}{-2.74}&4.61&27.27&22.66\\
\multicolumn{1}{c|}{}&\multicolumn{1}{c|}{50}&-3.43&-2.87&\multicolumn{1}{c|}{0.56}&-2.71&-3.07&\multicolumn{1}{c|}{-0.36}&-2.48&&\multicolumn{1}{c|}{}&-0.70&-2.35&-1.65&18.92&15.75&\multicolumn{1}{c|}{-3.17}&19.79&14.56&\multicolumn{1}{c|}{-5.24}&7.19&22.21&15.01\\
\multicolumn{1}{c|}{}&\multicolumn{1}{c|}{100}&-4.75&-5.34&\multicolumn{1}{c|}{-0.58}&-5.05&-5.29&\multicolumn{1}{c|}{-0.24}&-4.95&&\multicolumn{1}{c|}{}&0.34&-4.58&-4.92&-3.12&-2.89&\multicolumn{1}{c|}{0.23}&-3.39&-3.76&\multicolumn{1}{c|}{-0.37}&10.36&-3.26&{\ul -13.62}\\
\multicolumn{1}{c|}{}&\multicolumn{1}{c|}{200}&-2.59&-3.44&\multicolumn{1}{c|}{-0.84}&-3.60&-3.81&\multicolumn{1}{c|}{-0.21}&-3.11&&\multicolumn{1}{c|}{}&-0.92&-2.99&-2.07&-2.45&-2.60&\multicolumn{1}{c|}{-0.15}&-0.74&-3.64&\multicolumn{1}{c|}{-2.91}&2.71&-1.96&-4.67\\
\hline
\multicolumn{1}{c|}{\multirow{5}{*}{MEND}}&\multicolumn{1}{c|}{1}&0.86&\textbf{0.72}&\multicolumn{1}{c|}{-0.14}&0.07&-0.69&\multicolumn{1}{c|}{-0.76}&-0.15&-0.37&\multicolumn{1}{c|}{-0.22}&1.66&\textbf{-0.07}&{\ul -1.73}&1.29&&\multicolumn{1}{c|}{}&-0.11&&\multicolumn{1}{c|}{}&1.51&0.29&{\ul -1.22}\\
\multicolumn{1}{c|}{}&\multicolumn{1}{c|}{10}&-0.45&-1.26&\multicolumn{1}{c|}{-0.81}&-0.66&-1.60&\multicolumn{1}{c|}{-0.95}&-1.34&-1.77&\multicolumn{1}{c|}{-0.43}&1.73&\textbf{-0.36}&{\ul -2.08}&0.41&\textbf{1.65}&\multicolumn{1}{c|}{{\ul 1.24}}&2.80&\textbf{0.70}&\multicolumn{1}{c|}{-2.10}&8.87&3.79&{\ul -5.07}\\
\multicolumn{1}{c|}{}&\multicolumn{1}{c|}{\cancel{50}}&360.75&493.50&\multicolumn{1}{c|}{\cancel{132.75}}&427.41&450.42&\multicolumn{1}{c|}{\cancel{23.01}}&549.66&&\multicolumn{1}{c|}{}&137.39&455.15&\cancel{317.75}&89.14&76.42&\multicolumn{1}{c|}{\cancel{-12.72}}&71.07&70.72&\multicolumn{1}{c|}{\cancel{-0.36}}&45.04&75.22&\cancel{30.19}\\
\multicolumn{1}{c|}{}&\multicolumn{1}{c|}{\cancel{100}}&305.89&351.72&\multicolumn{1}{c|}{\cancel{45.83}}&362.27&229.10&\multicolumn{1}{c|}{\cancel{-133.17}}&338.70&&\multicolumn{1}{c|}{}&134.81&407.41&\cancel{272.61}&315.42&285.40&\multicolumn{1}{c|}{\cancel{-30.02}}&296.70&248.77&\multicolumn{1}{c|}{\cancel{-47.93}}&108.53&332.79&\cancel{224.26}\\
\multicolumn{1}{c|}{}&\multicolumn{1}{c|}{\cancel{200}}&361.28&401.57&\multicolumn{1}{c|}{\cancel{40.29}}&398.28&248.82&\multicolumn{1}{c|}{\cancel{-149.46}}&513.83&&\multicolumn{1}{c|}{}&170.13&459.61&\cancel{289.48}&428.39&390.63&\multicolumn{1}{c|}{\cancel{-37.76}}&340.96&280.78&\multicolumn{1}{c|}{\cancel{-60.17}}&150.13&442.50&\cancel{292.37}\\
\hline
\multicolumn{1}{c|}{\multirow{5}{*}{ROME}}&\multicolumn{1}{c|}{1}&-2.20&\textbf{1.05}&\multicolumn{1}{c|}{{\ul 3.24}}&-0.23&\textbf{}&\multicolumn{1}{c|}{-0.33}&-0.13&\textbf{4.05}&\multicolumn{1}{c|}{{\ul 4.18}}&4.69&\textbf{-0.96}&{\ul -5.65}&-1.19&&\multicolumn{1}{c|}{}&0.89&&\multicolumn{1}{c|}{}&6.33&\textbf{-0.06}&{\ul -6.39}\\
\multicolumn{1}{c|}{}&\multicolumn{1}{c|}{10}&1.88&\textbf{1.05}&\multicolumn{1}{c|}{-0.83}&0.10&-0.48&\multicolumn{1}{c|}{-0.58}&-0.27&\textbf{4.09}&\multicolumn{1}{c|}{{\ul 4.36}}&5.75&\textbf{-0.50}&{\ul -6.25}&3.55&\textbf{5.57}&\multicolumn{1}{c|}{{\ul 2.02}}&4.16&\textbf{7.43}&\multicolumn{1}{c|}{3.27}&6.73&2.00&{\ul -4.73}\\
\multicolumn{1}{c|}{}&\multicolumn{1}{c|}{\cancel{50}}&99.09&81.91&\multicolumn{1}{c|}{\cancel{-17.18}}&83.84&77.07&\multicolumn{1}{c|}{\cancel{-6.77}}&78.50&&\multicolumn{1}{c|}{}&64.81&90.04&\cancel{25.22}&921.70&980.84&\multicolumn{1}{c|}{\cancel{59.14}}&1016.98&1001.62&\multicolumn{1}{c|}{\cancel{-15.36}}&665.52&994.84&\cancel{329.32}\\
\multicolumn{1}{c|}{}&\multicolumn{1}{c|}{\cancel{100}}&112.31&88.60&\multicolumn{1}{c|}{\cancel{-23.71}}&92.36&85.94&\multicolumn{1}{c|}{\cancel{-6.41}}&84.03&&\multicolumn{1}{c|}{}&65.95&99.18&\cancel{33.23}&524.14&572.46&\multicolumn{1}{c|}{\cancel{48.33}}&465.61&570.95&\multicolumn{1}{c|}{\cancel{105.34}}&244.14&458.92&\cancel{214.78}\\
\multicolumn{1}{c|}{}&\multicolumn{1}{c|}{\cancel{200}}&226.50&204.29&\multicolumn{1}{c|}{\cancel{-22.21}}&201.34&197.18&\multicolumn{1}{c|}{\cancel{-4.16}}&248.73&&\multicolumn{1}{c|}{}&229.24&230.52&\cancel{1.28}&386.17&359.52&\multicolumn{1}{c|}{\cancel{-26.65}}&461.55&346.48&\multicolumn{1}{c|}{\cancel{-115.08}}&244.37&415.69&\cancel{171.33}\\
\hline
\multicolumn{1}{c|}{\multirow{5}{*}{MEMIT}}&\multicolumn{1}{c|}{1}&0.32&\textbf{0.59}&\multicolumn{1}{c|}{{\ul 0.27}}&0.62&\textbf{0.48}&\multicolumn{1}{c|}{-0.14}&0.32&\textbf{0.61}&\multicolumn{1}{c|}{{\ul 0.28}}&-4.10&0.34&4.44&-2.73&&\multicolumn{1}{c|}{}&-0.17&&\multicolumn{1}{c|}{}&2.31&\textbf{-0.32}&{\ul -2.63}\\
\multicolumn{1}{c|}{}&\multicolumn{1}{c|}{10}&-0.82&-0.22&\multicolumn{1}{c|}{{\ul 0.60}}&0.41&\textbf{0.69}&\multicolumn{1}{c|}{{\ul 0.28}}&0.01&-0.22&\multicolumn{1}{c|}{-0.23}&-4.96&0.19&5.14&-0.09&\textbf{1.33}&\multicolumn{1}{c|}{{\ul 1.42}}&-0.26&-0.21&\multicolumn{1}{c|}{{\ul 0.05}}&2.13&\textbf{-1.47}&{\ul -3.60}\\
\multicolumn{1}{c|}{}&\multicolumn{1}{c|}{50}&-0.65&-0.19&\multicolumn{1}{c|}{{\ul 0.46}}&-0.75&-0.34&\multicolumn{1}{c|}{{\ul 0.41}}&-0.32&&\multicolumn{1}{c|}{}&2.70&\textbf{-0.79}&{\ul -3.49}&-0.38&\textbf{0.85}&\multicolumn{1}{c|}{{\ul 1.23}}&-0.20&-0.52&\multicolumn{1}{c|}{-0.32}&3.26&\textbf{-1.06}&{\ul -4.31}\\
\multicolumn{1}{c|}{}&\multicolumn{1}{c|}{100}&-0.87&\textbf{0.08}&\multicolumn{1}{c|}{{\ul 0.95}}&-0.68&-0.12&\multicolumn{1}{c|}{{\ul 0.56}}&-0.13&&\multicolumn{1}{c|}{}&3.30&\textbf{-0.89}&{\ul -4.19}&0.14&\textbf{1.15}&\multicolumn{1}{c|}{{\ul 1.02}}&-0.42&\textbf{0.64}&\multicolumn{1}{c|}{{\ul 1.06}}&2.65&\textbf{-0.79}&{\ul -3.44}\\
\multicolumn{1}{c|}{}&\multicolumn{1}{c|}{200}&-0.66&\textbf{1.18}&\multicolumn{1}{c|}{{\ul 1.83}}&0.34&\textbf{0.31}&\multicolumn{1}{c|}{-0.03}&0.04&&\multicolumn{1}{c|}{}&2.61&\textbf{-0.80}&{\ul -3.41}&1.08&\textbf{1.54}&\multicolumn{1}{c|}{{\ul 0.46}}&1.42&\textbf{0.45}&\multicolumn{1}{c|}{-0.97}&4.05&0.86&{\ul -3.19}\\

\hline
\multicolumn{1}{c|}{\multirow{5}{*}{EREN}}&\multicolumn{1}{c|}{1}&
0.249&\textbf{0.353}&\multicolumn{1}{c|}{{\ul0.104}}&
0.496&\textbf{1.564}&\multicolumn{1}{c|}{{\ul1.068}}&
0&\textbf{1.259}&\multicolumn{1}{c|}{{\ul1.259}}&
1.654&2.941&\multicolumn{1}{c|}{{\ul1.287}}&
3.482&&\multicolumn{1}{c|}{}&
2.144&&\multicolumn{1}{c|}{}&
0&4.125&\multicolumn{1}{c}{{\ul 4.125}}\\

\multicolumn{1}{c|}{}&\multicolumn{1}{c|}{10}&
12.248&\textbf{51.441}&\multicolumn{1}{c|}{{\ul39.193}}&
3.458&\textbf{27.158}&\multicolumn{1}{c|}{{\ul23.700}}&
0.843&\textbf{5.186}&\multicolumn{1}{c|}{{\ul4.343}}&
0.159&1.588&\multicolumn{1}{c|}{{\ul1.429}}&
19.683&&\multicolumn{1}{c|}{}&
14.638&&\multicolumn{1}{c|}{}&
4.683&2.582&\multicolumn{1}{c}{{-2.101}}\\

\multicolumn{1}{c|}{}&\multicolumn{1}{c|}{50}&
25.382&\textbf{49.159}&\multicolumn{1}{c|}{{\ul23.777}}&
11.490&\textbf{21.514}&\multicolumn{1}{c|}{{\ul10.024}}&
12.385&5.693&\multicolumn{1}{c|}{-6.692}&
5.173&2.581&\multicolumn{1}{c|}{-2.592}&
26.395&\textbf{51.502}&\multicolumn{1}{c|}{{\ul25.107}}&
13.582&\textbf{18.122}&\multicolumn{1}{c|}{{\ul4.540}}&
4.891&21.481&\multicolumn{1}{c}{{-16.59}}\\

\multicolumn{1}{c|}{}&\multicolumn{1}{c|}{100}&
71.491&\textbf{152.252}&\multicolumn{1}{c|}{{\ul80.761}}&
114.529&\textbf{118.292}&\multicolumn{1}{c|}{{\ul3.763}}&
125.713&113.965&\multicolumn{1}{c|}{-11.748}&
15.486&2.415&\multicolumn{1}{c|}{-13.071}&
59.496&\textbf{159.797}&\multicolumn{1}{c|}{{\ul100.301}}&
23.853&\textbf{142.592}&\multicolumn{1}{c|}{{\ul118.739}}&
124.547&119.592&\multicolumn{1}{c}{{-4.955}}\\

\multicolumn{1}{c|}{}&\multicolumn{1}{c|}{200}&
125.951&\textbf{217.592}&\multicolumn{1}{c|}{{\ul91.641}}&
138.569&\textbf{141.493}&\multicolumn{1}{c|}{{\ul2.924}}&
135.471&111.548&\multicolumn{1}{c|}{-23.923}&
217.468&118.582&\multicolumn{1}{c|}{-98.886}&
189.491&\textbf{235.962}&\multicolumn{1}{c|}{{\ul46.471}}&
115.394&\textbf{168.633}&\multicolumn{1}{c|}{{\ul 53.239}}&
171.502&85.348&\multicolumn{1}{c}{{-86.154}}\\

\hline

\multicolumn{1}{c|}{\multirow{5}{*}{SERAC}}&\multicolumn{1}{c|}{1}&
0&0&\multicolumn{1}{c|}{0}&
0&\textbf{0}&\multicolumn{1}{c|}{0}&
0&\textbf{0}&\multicolumn{1}{c|}{0}&
0&0&\multicolumn{1}{c|}{0}&
0&&\multicolumn{1}{c|}{}&
0&&\multicolumn{1}{c|}{}&
0&0&\multicolumn{1}{c}{0}\\

\multicolumn{1}{c|}{}&\multicolumn{1}{c|}{10}&
1.677&\textbf{2.481}&\multicolumn{1}{c|}{{\ul0.804}}&
2.314&0.512&\multicolumn{1}{c|}{-1.802}&
0.487&0.193&\multicolumn{1}{c|}{-0.294}&
2.174&\textbf{2.885}&\multicolumn{1}{c|}{{\ul0.711}}&
3.795&&\multicolumn{1}{c|}{}&
1.483&&\multicolumn{1}{c|}{}&
1.583&1.195&\multicolumn{1}{c}{{-0.388}}\\

\multicolumn{1}{c|}{}&\multicolumn{1}{c|}{50}&
15.391&\textbf{21.460}&\multicolumn{1}{c|}{{\ul6.069}}&
18.631&17.623&\multicolumn{1}{c|}{-1.008}&
14.582&13.582&\multicolumn{1}{c|}{-1.000}&
15.0284&3.991&\multicolumn{1}{c|}{-11.037}&
25.493&\textbf{25.781}&\multicolumn{1}{c|}{{\ul0.288}}&
28.974&20.581&\multicolumn{1}{c|}{-8.393}&
14.592&15.639&\multicolumn{1}{c}{{\ul 1.047}}\\

\multicolumn{1}{c|}{}&\multicolumn{1}{c|}{100}&
128.682&\textbf{137.952}&\multicolumn{1}{c|}{{\ul9.270}}&
85.162&74.592&\multicolumn{1}{c|}{-10.570}&
136.581&55.502&\multicolumn{1}{c|}{-81.079}&
145.610&87.910&\multicolumn{1}{c|}{-57.700}&
158.542&\textbf{207.539}&\multicolumn{1}{c|}{{\ul48.997}}&
124.632&64.623&\multicolumn{1}{c|}{-60.009}&
169.531&88.593&\multicolumn{1}{c}{-80.938}\\

\multicolumn{1}{c|}{}&\multicolumn{1}{c|}{200}&
89.246&\textbf{174.642}&\multicolumn{1}{c|}{{\ul85.396}}&
115.223&\textbf{115.245}&\multicolumn{1}{c|}{{\ul0.022}}&
119.128&85.602&\multicolumn{1}{c|}{-33.526}&
151.963&96.942&\multicolumn{1}{c|}{-55.021}&
137.532&\textbf{185.938}&\multicolumn{1}{c|}{{\ul48.406}}&
145.631&79.105&\multicolumn{1}{c|}{-66.526}&
96.325&85.203&\multicolumn{1}{c}{-11.122}\\

\hline
\multicolumn{1}{c|}{\multirow{6}{*}{ICE}}&\multicolumn{1}{c|}{1}&2.166&\textbf{6.353}&\multicolumn{1}{c|}{{\ul 4.187}}&2.525&\textbf{1.589}&\multicolumn{1}{c|}{-0.936}&2.588&\textbf{3.963}&\multicolumn{1}{c|}{{\ul 1.375}}&232.538&2.22&{\ul -230.318}&0.202&&\multicolumn{1}{c|}{}&3.544&&\multicolumn{1}{c|}{}&27.124&4.343&{\ul-22.781}\\
\multicolumn{1}{c|}{}&\multicolumn{1}{c|}{2}&4.976&\textbf{6.325}&\multicolumn{1}{c|}{{\ul 1.349}}&2.997&\textbf{2.883}&\multicolumn{1}{c|}{-0.114}&3.239&\textbf{3.713}&\multicolumn{1}{c|}{{\ul0.474}}&47.943&2.16&{\ul-45.783}&2.871&&\multicolumn{1}{c|}{}&2.457&&\multicolumn{1}{c|}{}&9.056&1.04&{\ul-8.016}\\
\multicolumn{1}{c|}{}&\multicolumn{1}{c|}{3}&3.79&\textbf{3.476}&\multicolumn{1}{c|}{-0.314}&1.77&\textbf{1.616}&\multicolumn{1}{c|}{-0.154}&2.461&\textbf{4.852}&\multicolumn{1}{c|}{{\ul2.391}}&7.223&1.59&{\ul-5.633}&1.138&\textbf{3.286}&\multicolumn{1}{c|}{{\ul2.148}}&3.003&\textbf{1.662}&\multicolumn{1}{c|}{-1.342}&0.453&1.753&1.3\\
\multicolumn{1}{c|}{}&\multicolumn{1}{c|}{5}&1.171&\textbf{2.221}&\multicolumn{1}{c|}{{\ul1.05}}&0.762&\textbf{1.738}&\multicolumn{1}{c|}{{\ul0.976}}&2.522&\textbf{2.478}&\multicolumn{1}{c|}{-0.044}&10.261&0.989&{\ul-9.272}&4.447&\textbf{6.518}&\multicolumn{1}{c|}{{\ul2.071}}&4.413&\textbf{5.047}&\multicolumn{1}{c|}{{\ul0.634}}&11.791&2.425&{\ul-9.366}\\
\multicolumn{1}{c|}{}&\multicolumn{1}{c|}{8}&3.491&\textbf{3.238}&\multicolumn{1}{c|}{-0.253}&1.329&\textbf{2.18}&\multicolumn{1}{c|}{{\ul0.851}}&8.009&\textbf{4.195}&\multicolumn{1}{c|}{-3.814}&7.224&4.694&{\ul-2.53}&2.118&\textbf{5.011}&\multicolumn{1}{c|}{{\ul2.893}}&3.096&\textbf{2.857}&\multicolumn{1}{c|}{-0.238}&3.297&1.866&{\ul-1.431}\\
\multicolumn{1}{c|}{}&\multicolumn{1}{c|}{10}&2.741&\textbf{12.489}&\multicolumn{1}{c|}{{\ul9.748}}&1.279&\textbf{2.611}&\multicolumn{1}{c|}{{\ul1.332}}&8.95&\textbf{3.577}&\multicolumn{1}{c|}{-5.373}&5.371&1.214&{\ul-4.157}&4.408&\textbf{6.058}&\multicolumn{1}{c|}{{\ul1.65}}&4.684&\textbf{5.756}&\multicolumn{1}{c|}{{\ul1.072}}&5.166&3.144&{\ul-2.022}\\
\hline
\multicolumn{1}{c|}{\multirow{5}{*}{SIR\_top5}}&\multicolumn{1}{c|}{1}&-0.067&\textbf{2.863}&\multicolumn{1}{c|}{{\ul2.93}}&-0.499&-0.94&\multicolumn{1}{c|}{-0.441}&-0.054&-0.869&\multicolumn{1}{c|}{-0.815}&3.261&\textbf{-1.399}&{\ul-4.66}&-3.631&&\multicolumn{1}{c|}{}&-0.213&&\multicolumn{1}{c|}{}&2.617&\textbf{-0.35}&{\ul-2.967}\\
\multicolumn{1}{c|}{}&\multicolumn{1}{c|}{10}&-0.862&\textbf{2.349}&\multicolumn{1}{c|}{{\ul3.211}}&-0.518&\textbf{-0.501}&\multicolumn{1}{c|}{{\ul0.017}}&-0.009&-1.021&\multicolumn{1}{c|}{-1.012}&2.611&\textbf{-1.34}&{\ul-3.951}&-0.318&\textbf{0.905}&\multicolumn{1}{c|}{{\ul1.223}}&-0.451&\textbf{0.157}&\multicolumn{1}{c|}{{\ul0.608}}&1.811&\textbf{-1.539}&{\ul-3.350}\\
\multicolumn{1}{c|}{}&\multicolumn{1}{c|}{50}&-0.322&-0.916&\multicolumn{1}{c|}{-0.594}&-0.727&-0.073&\multicolumn{1}{c|}{{\ul0.654}}&-0.083&&\multicolumn{1}{c|}{}&2.511&\textbf{-1.418}&{\ul-3.929}&-0.634&\textbf{0.682}&\multicolumn{1}{c|}{{\ul1.316}}&-0.239&-0.699&\multicolumn{1}{c|}{-0.460}&2.064&\textbf{-1.098}&{\ul-3.162}\\
\multicolumn{1}{c|}{}&\multicolumn{1}{c|}{100}&-0.982&\textbf{0.828}&\multicolumn{1}{c|}{{\ul1.81}}&-0.586&-0.092&\multicolumn{1}{c|}{{\ul0.494}}&-0.087&&\multicolumn{1}{c|}{}&2.796&\textbf{-1.343}&{\ul-4.139}&0.066&1.132&\multicolumn{1}{c|}{{\ul1.066}}&-0.154&-0.522&\multicolumn{1}{c|}{-0.368}&1.968&\textbf{-0.739}&{\ul-2.707}\\
\multicolumn{1}{c|}{}&\multicolumn{1}{c|}{200}&-0.804&\textbf{0.979}&\multicolumn{1}{c|}{{\ul1.783}}&0.331&\textbf{0.544}&\multicolumn{1}{c|}{{\ul0.213}}&0.036&&\multicolumn{1}{c|}{}&2.219&\textbf{-0.815}&{\ul-3.034}&0.318&\textbf{1.417}&\multicolumn{1}{c|}{{\ul1.099}}&-0.329&-0.527&\multicolumn{1}{c|}{-0.198}&1.241&\textbf{-0.476}&{\ul-1.717}\\\hline
\multicolumn{1}{c|}{\multirow{5}{*}{SIR\_top10}}&\multicolumn{1}{c|}{1}&-0.106&\textbf{2.148}&\multicolumn{1}{c|}{{\ul2.254}}&-0.706&-0.933&\multicolumn{1}{c|}{-0.227}&-0.21&-0.61&\multicolumn{1}{c|}{-0.400}&2.544&\textbf{-1.394}&{\ul-3.938}&-1.365&&\multicolumn{1}{c|}{}&-0.117&&\multicolumn{1}{c|}{}&2.048&\textbf{-0.384}&{\ul-2.432}\\
\multicolumn{1}{c|}{}&\multicolumn{1}{c|}{10}&-0.798&\textbf{2.473}&\multicolumn{1}{c|}{{\ul3.271}}&-0.565&-0.51&\multicolumn{1}{c|}{{\ul0.055}}&0.309&-0.841&\multicolumn{1}{c|}{-1.151}&3.168&\textbf{-1.226}&{\ul-4.394}&-0.571&\textbf{0.829}&\multicolumn{1}{c|}{{\ul1.4}}&-0.496&-0.017&\multicolumn{1}{c|}{{\ul0.479}}&1.563&\textbf{-1.549}&{\ul-3.112}\\
\multicolumn{1}{c|}{}&\multicolumn{1}{c|}{50}&-0.406&\textbf{0.578}&\multicolumn{1}{c|}{{\ul0.984}}&-0.939&-0.065&\multicolumn{1}{c|}{{\ul0.874}}&-0.222&&\multicolumn{1}{c|}{}&1.29&\textbf{-1.552}&{\ul-2.842}&-0.57&\textbf{0.565}&\multicolumn{1}{c|}{{\ul1.135}}&-0.281&-1.216&\multicolumn{1}{c|}{-0.935}&2.489&\textbf{-0.985}&{\ul-3.474}\\
\multicolumn{1}{c|}{}&\multicolumn{1}{c|}{100}&-0.703&\textbf{0.741}&\multicolumn{1}{c|}{{\ul1.444}}&-0.705&-0.284&\multicolumn{1}{c|}{{\ul0.421}}&-0.133&&\multicolumn{1}{c|}{}&1.47&\textbf{-1.304}&{\ul-2.774}&-0.078&\textbf{0.876}&\multicolumn{1}{c|}{{\ul0.954}}&-0.084&\textbf{0.012}&\multicolumn{1}{c|}{{\ul0.096}}&1.404&\textbf{-0.769}&{\ul-2.173}\\
\multicolumn{1}{c|}{}&\multicolumn{1}{c|}{200}&-0.838&\textbf{0.947}&\multicolumn{1}{c|}{{\ul1.785}}&0.339&\textbf{0.481}&\multicolumn{1}{c|}{{\ul0.142}}&0.1911&&\multicolumn{1}{c|}{}&1.772&\textbf{-0.728}&{\ul-2.5}&0.234&\textbf{1.421}&\multicolumn{1}{c|}{{\ul1.187}}&-0.054&-0.31&\multicolumn{1}{c|}{-0.256}&1.528&\textbf{-0.575}&{\ul-2.103}\\\hline
\end{tabular}
}
\vspace{-2mm}
\caption{Comparative analysis of perplexity changes. 
The first row categorizes the distribution of edits, and the second row indicates the distances between affected and edited triplets, with ``inf'' signifying no connectivity. 
``Vanilla'' denotes the change in perplexity on the vanilla knowledge graph before and after edits, whereas ``GIE'' signifies the change in perplexity following the application of GIE.
The ``Diff'' column is obtained by subtracting ``Vanilla'' from ``GIE''. 
Editing methods are specified in the leftmost column, while the adjacent column enumerates the number of edits applied. 
The slashed values indicate the method's inability to accommodate the quantity of edits. 
Underlined values signify that the ripple effect in hidden space is more obvious than the other two variants.
Bolded values indicate the presence of a ripple effect in hidden space, which is successfully discerned via GIE.
}
\label{tab:experiment}
\end{table*}